\documentclass[11pt]{article}

\usepackage[final]{acl}

\usepackage{times}
\usepackage{latexsym}

\usepackage[T1]{fontenc}

\usepackage[utf8]{inputenc}

\usepackage{microtype}

\usepackage{inconsolata}

\usepackage{graphicx}

\definecolor{todo}{rgb}{1,0.5,0}
\definecolor{rev}{rgb}{0.8,0,0.6}

\newcommand{\ext}[1]{\textcolor{orange}{}} 
\usepackage{enumitem}
\usepackage{epigraph}
\usepackage{amssymb}
\usepackage{tabularx,booktabs,makecell, multirow}
\usepackage{listings}
\usepackage{fancyvrb}
\usepackage{wasysym}
\usepackage{float}

\usepackage{enumitem}
\usepackage{array}
\newcolumntype{C}[1]{>{\centering\arraybackslash}p{#1}}
\newcommand\blfootnote[1]{%
  \begingroup
  \renewcommand\thefootnote{}\footnote{#1}%
  \addtocounter{footnote}{-1}%
  \endgroup
}

\title{Is Peer Review Really in Decline? \\ Analyzing Review Quality across Venues and Time} 
\author{
Ilia Kuznetsov\textsuperscript{*1}, Rohan Nayak\textsuperscript{*1}, Alla Rozovskaya\textsuperscript{2}, Iryna Gurevych\textsuperscript{1}
\\
        \textsuperscript{1}Ubiquitous Knowledge Processing Lab (UKP Lab), \\ Department of Computer Science, Technical University of Darmstadt\\ and National Research Center for Applied Cybersecurity ATHENE\\
\textsuperscript{2}Department of Computer Science at Queens College, \\ City University of New York (CUNY)\\
\href{https://www.ukp.tu-darmstadt.de}{www.ukp.tu-darmstadt.de}
}

\begin{document}
\maketitle
\blfootnote{\hspace*{-0.112cm}\textsuperscript{*}Equal contribution. Rohan Nayak performed the work during an internship at UKP Lab, now at Intelligent Language Systems Lab, FAU Erlangen-Nürnberg.}
\begin{abstract}
Peer review is at the heart of modern science. As submission numbers rise and research communities grow, the decline in review quality is a popular narrative and a common concern. Yet, is it true? Review quality is difficult to measure, and the ongoing evolution of reviewing practices makes it difficult to compare reviews across venues and time. To address this, we introduce a new framework for the comparative study of review quality and apply it to major AI and machine learning conferences. We propose a multi-dimensional schema to serve as a proxy for quantifying review quality in terms of its utility to editors and authors. The schema is coupled with LLM-based and lightweight measurements, which we validate against a novel test set of human-labeled review quality ratings. Contradicting the popular narrative, our proxy measurements reveal no consistent decline in median review quality across venues and years. We propose alternative explanations, and outline recommendations to facilitate future empirical studies of review quality.\footnote{Code and data available on \href{https://github.com/UKPLab/emnlp2026-review-quality-estimation}{GitHub}.}
\end{abstract}

\section{Introduction}
\label{sec:intro}
%

Peer review is central to the modern academic practice. 
It helps establish the epistemic status of research findings (“Is this science?”) and prioritize research outputs. The results of peer review determine funding decisions, shape public opinion, and inform policy \cite{Latona2024-qk}. Peer review is deployed globally and by most scientific fields, across a wide range of publication and presentation formats. Computer science and machine learning conferences are no exception and use peer review as the primary quality control mechanism.

Despite its merits, peer review is time-consuming \cite{GSPR} and prone to bias, error and randomness \cite{lee, Rogers2020-bo, beygelzimer2023machinelearningreviewprocess}. Acceptance decisions at major computer science conferences are accompanied by frustration on social media,\footnote{Paraphrasing a Reddit user, \textit{“You spend a week working like a dog, running pointless experiments to make reviewers happy so that they increase your score.”} \cite{reddit:neurips2025_rebuttals}} and the concerns about the declining review quality are widespread \cite{Kim2025-ic, Adam2025-ix}. There are indeed reasons to suspect that peer review is in decline: qualified reviewers are scarce, increasing specialization requires deeper expertise, reviewers and editors suffer high workloads, and irresponsible use of large language models (LLMs) creates new risks \cite{Latona2024-qk}. Yet, there are also reasons to believe that review quality didn't decline after all. Conferences refine their reviewing workflows, the pool of experts grows with each completed reviewing campaign, and ethical use of AI assistance can support reviewers in their work. All in all, the evidence for either side remains largely anecdotal \cite{Adam2025-ix}, and the true state of peer review is unknown.

\begin{figure*}
    \centering
    \includegraphics[width=0.9\linewidth]{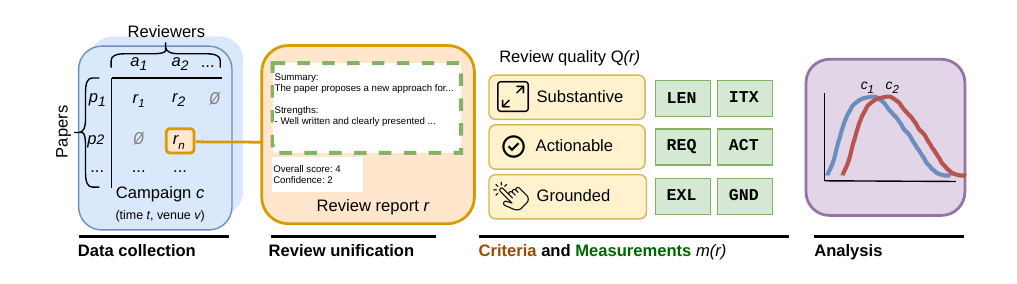}
    \caption{Framework overview. To study the evolution of peer review quality, we collect data across venues and time, bring the reviews into a unified format, propose six measurements across three general review quality criteria, and aggregate measurement results for analysis. Formally, a review report $r$ is written for a paper $p$ by a reviewer $a$, and originates from a reviewing campaign $c$ that takes place as part of a venue $v$ at time $t$, e.g. ICLR-2018. Reviews differ in terms of their quality $Q(r)$, approximated through a range of measurements $m(r) \in M$ along a set of quality criteria. Aggregating review-level measurements allows comparing review quality across campaigns.
    }
    \label{fig:framework}
\end{figure*}

To study changes in review quality, evaluate new tools and policies, and detect low-quality reviews, an objective and scalable \emph{measure of review quality} is needed. Conceptualizing such measure is difficult  (Section \ref{sec:relwork}): review quality is hard to formalize, human judgment is prone to bias, and extrinsic measurements such as citation counts are easily distorted. Practical challenges further complicate the picture: the data is scarce, biased and noisy, and the diversity and evolution of reviewing workflows makes consistent application of review quality measures difficult. While the need to quantify review quality is much acknowledged, a general framework for multi-dimensional analysis of review quality at scale is missing. To address this gap, our study makes the following \textbf{contributions}: 

\begin{itemize}[noitemsep, topsep=0pt]
\item We propose the first holistic automated \emph{framework} to study review quality at scale (Fig. \ref{fig:framework}), tackling the issues of review unification, quality measurement and analysis.
\item We propose a new \emph{method for review quality measurement} across three dimensions -- substantiveness, actionability, and grounding -- that approximate review value to editors and authors. For each, we implement two \emph{automated measurements}: a lightweight (simple and efficient), and an LLM-based.
\item We develop and release new \textit{guidelines} for human evaluation of review quality, validated through inter-annotator agreement, coupled with a new manually labeled \textit{evaluation set}.
\item We \emph{validate} our measurements against human ratings on our dataset and a further, independent third-party dataset.
\item We use our measurements to study the evolution of review quality at multiple AI and machine learning conferences.
\end{itemize}

\noindent We find that LLMs can help unify review structure across venues and time. The proposed measurements of review quality are correlated but complementary, and lightweight measurements can approximate LLM-based measurements at low cost. A moderate-strong agreement with human ratings demonstrates the promise of our approach. Contrary to the widespread perception, our proxy measurements reveal no consistent decline in the median review quality at computer science conferences. Yet, the question proves nuanced, and we provide five hypotheses coupled with recommendations to researchers, conference organizers and community leaders to move this work forward (Section \ref{sec:discussion}).


\section{Scope of the study}
\label{sec:scope}

Peer review is a multi-stage process spanning from the call for papers to acceptance decisions, and a failure at any stage can undermine the goals of peer review \cite{kuznetsov2024natural}. Review reports are central to this process and are the most consistently documented reviewing artifact \cite{kang-etal-2018-dataset, lin2023moprd, dycke-etal-2023-nlpeer}. Thus, our study focuses on the \emph{quality of review reports}. Peer review involves stakeholders with distinct and often contradictory goals: for example, authors seek acceptance and constructive feedback, while editors aim to select sound, impactful and thematically fitting papers. We focus on modeling review quality in terms of utility for both \emph{editors and authors}: this guides our decisions in both selecting the measurements (Section \ref{sec:measurements}) and designing the human evaluation (Section \ref{sec:humeval}). In terms of domain, we focus on \textit{conferences in computer science} that make their peer reviews available: The International Conference on Learning Representations (ICLR), The Conference on Neural Information Processing Systems (NeurIPS), and ACL Rolling Review (ARR). These venues employ double-blind pre-publication review \cite{wiley_peer_review_types} and experience rapid submission growth, serving as a representative sample of peer reviewing in the community. Yet, the framework itself is general and can be adapted to other fields and publication workflows (see \nameref{sec:limitations}). 

\section{Related work}
\label{sec:relwork}

Approaches to \textbf{measuring review quality} fall into three categories. \emph{Survey-based approaches} rely on human judgment, from flagging problematic reviews \cite{Rogers2023-kh} to structured measurement instruments \cite{Rooyen1999}. While insightful, human judgment is prone to bias \cite{Goldberg2023-rk}, does not scale, and is difficult to apply retroactively, making it a poor fit for comparative analysis. \emph{Extrinsic approaches} evaluate the downstream effectiveness of peer review, e.g. through the use of future paper citations \cite{plank2019citetracked, Maddi2024-fw}. While useful for measuring the impact of peer review as a system, extrinsic measurements need time to play out, and are affected by external factors such as topic popularity or advertisement efforts \cite{tahamtan, luc2021does}. An alternative, \emph{Formal approach} to review quality measurement estimates review quality by calculating indicators based on review contents. This allows objective measurement that can be automated. 

While multi-dimensional schemata of review quality do exist (e.g. \citet{Rooyen1999}), to the best of our knowledge, we are the first to propose a \textit{holistic}, \textit{automated} framework for formal analysis of review quality and apply it \textit{at scale}. A concurrent work by \citet{rotrev} proposes a set of formal review quality measurements, but does not tackle the issues of review standardization and measurement aggregation, preventing cross-venue and cross-temporal analysis. We use their annotations along with our own in-house ratings to validate our measurements in Section \ref{sec:humeval}.

Formal approaches benefit from the rich prior work on \textbf{NLP for peer review analysis}. Existing work includes score prediction \cite{kang-etal-2018-dataset}, discourse and revision analysis \cite{hua-etal-2019-argument, re3}, 
lazy thinking detection \cite{purkayastha-etal-2025-lazyreview}, among others.
While these works typically focus on \textit{individual} dimensions of review quality and optimize performance on selected datasets, we are the first to explore \textit{interactions} between different measurements and address the challenges of applying them across \textit{diverse} reviewing data. Furthermore, in addition to the LLM-based predictors, we investigate simple measurements of peer review quality, and show that they can serve as approximates for more resource-intensive methods. 

A common assumption in the literature is that review reports follow a standard \textbf{structure}, e.g. contain certain sections or are organized in bullet points \cite{rnr, dycke-etal-2023-overview, sadallah-etal-2025-good}. Yet, in practice, report structures and practices evolve over time, making such approaches not directly suitable for large-scale, cross-temporal analysis. While related prior work \cite{yu-etal-2024-automated} focuses on the task of aggregating reviews for the same paper evaluated automatically, we propose a new approach to review structure unification \emph{across venues and time}, coupled with an in-depth analysis along two dimensions: content preservation and section assignment.

Finally, measuring quality of human-written reviews is related, but distinct from \textbf{machine-generated review evaluation}. Prior NLP work on review generation partially or fully relies on human reference \cite{yu-etal-2024-automated, idahl-ahmadi-2025-openreviewer, zhu-etal-2025-deepreview, weng2025cycleresearcherimprovingautomatedresearch}. Yet, human-authored reviews themselves might be of higher or lower quality, and we lack the true gold standard to evaluate human reviews using a similar approach. Our work tackles this challenge by proposing a set of \textit{reference-free} metrics of review quality. While our focus here is on human reviews and cross-temporal analysis, the framework can be used to evaluate and compare machine-generated reviews, as well as to select high-quality human reviews for developing and evaluating such systems.




\section{Review unification}
\label{sec:unification}

\begin{figure}[t]
    \centering
    \includegraphics[width=.9\linewidth]{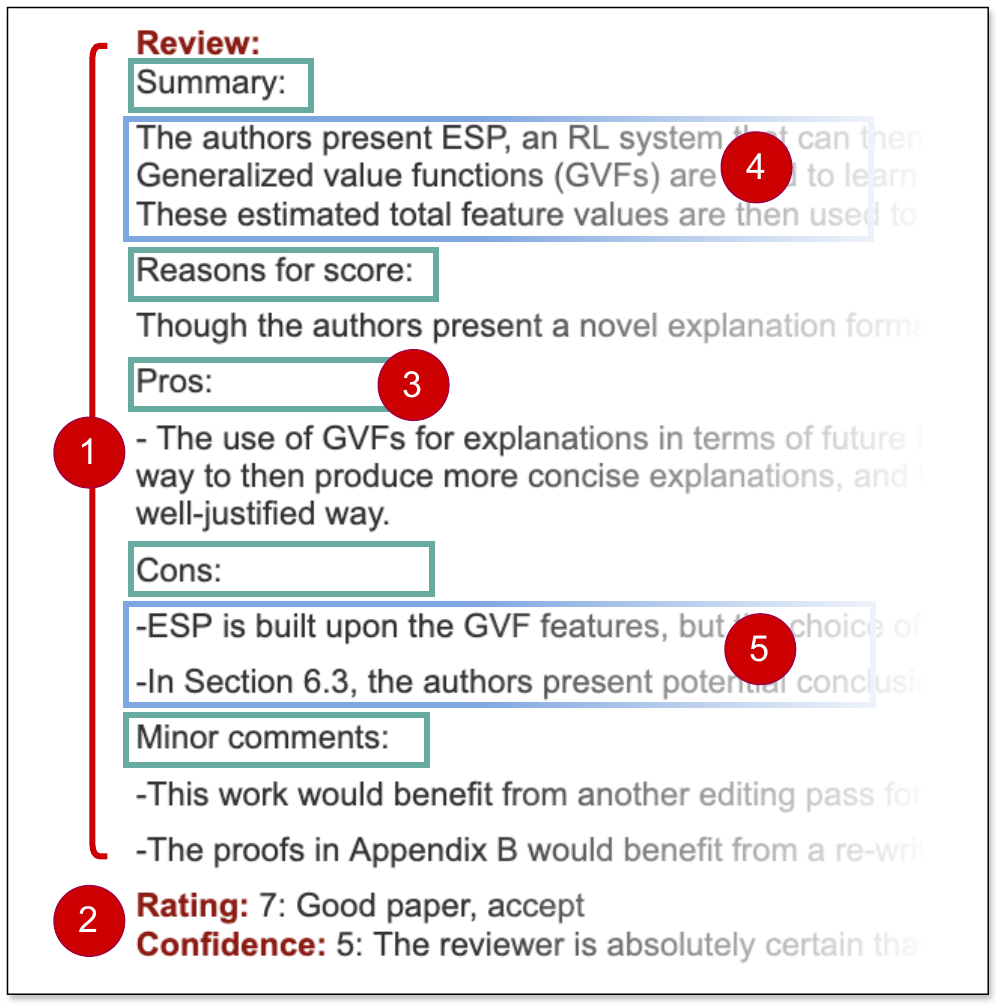}
    \caption{Structure of a review report (Section \ref{sec:rsd}).}
    \label{fig:rev-example}
\end{figure}

\subsection{Diversity of review structures}
\label{sec:rsd}


Peer reviews are semi-structured documents (Figure \ref{fig:rev-example}) typically composed of free-text (1), numerical (2), and other fields. Our analysis of review forms historically deployed at ICLR, NeurIPS and ARR (Table A.\ref{tab:form-fields}) shows that review structure can vary substantially. For example, while ICLR 2018-2021 used a single “Review” field for review text, from ICLR-2023 on, a structured form is used, which again is different from review forms at NeurIPS and ARR. Differences in review forms can affect measurements of review quality. In addition, individual reviewers can further organize their reports. Even within a single review field, the contents can be grouped into custom sections (Figure \ref{fig:rev-example}.3), presented as a narrative (4), or as lists and bullet points (5). To enable fair comparison, we need to cast peer reviews into a common representation.


\subsection{Flattening and itemization}

\begin{figure}[t]
    \centering
    \includegraphics[width=0.85\linewidth]{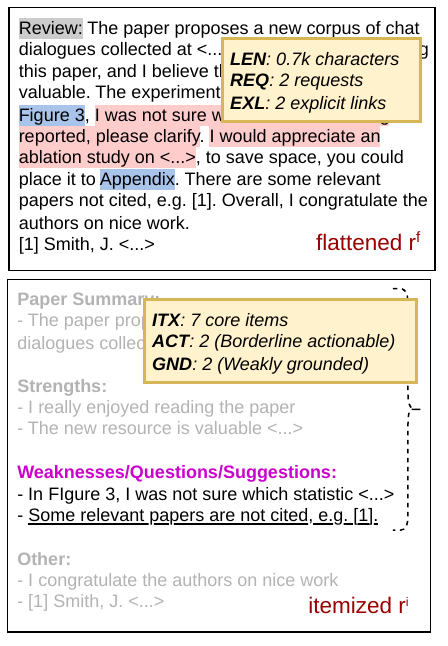}
    \caption{Measurements examples. \texttt{LEN}, \texttt{REQ} and \texttt{EXL} are calculated on the flattened review (top). \texttt{ITX} is calculated on the Summary, Strengths and Weaknesses+ sections of the itemized review (bottom). \texttt{ACT} and \texttt{GND} averaged over the Weaknesses+ section.}
    \label{fig:measurement_example}
    \vspace{-2mm}
\end{figure}

We address this variation in two ways. To unify reviews produced using differently structured forms, we \textit{flatten} the reviews by concatenating all the text fields and section headers from a review form into a continuous text (Figure \ref{fig:measurement_example}, top), $r \rightarrow r^f$, while discarding all non-textual fields. The result of this transformation is a \emph{flattened review} $r^f$. 
Addressing the differences in individual reporting styles is more challenging. Bullet-point style reporting is widely used at computer science conferences, and is a convenient unit of analysis, as it separates the review into thematically coherent segments which can be extracted via pattern matching \cite{yu-etal-2024-automated, sadallah-etal-2025-good}. Yet, not all reviews are composed this way, preventing the application of segment-level analysis across campaigns. To mitigate this, we propose the task of \emph{review itemization}, consisting of two subtasks. First, a review $r^f$ is transformed into a set of self-contained bullet point-style \emph{items} $i_k \in I(r^f)$. Second, each item is assigned to one of pre-defined \emph{sections} inspired by existing reviewing forms and the work on peer review pragmatics \cite{rnr}: \texttt{Summary}, \texttt{Strengths}, \texttt{Weaknesses+} and \texttt{Other}, where \texttt{Weaknesses+} aggregates all author-facing comments including weaknesses, questions, suggestions and comments. Importantly, this transformation must preserve the original content to maximum extent and should not introduce any new content. The result of this transformation is the \emph{itemized review} $r^i$ (Figure \ref{fig:measurement_example}, bottom).

\subsection{Implementation and evaluation}
\label{sec:itemeval}

Due to the lack of ground truth data on review unification \cite{yu-etal-2024-automated}, we use a zero-shot approach based on GPT-5.2.\footnote{Our attempts at using an open model were not successful, but we do not believe this to be a fundamental limitation; see \nameref{sec:ethics}.} Given a flattened review, we instruct the model to transform it into a series of bullet points assigned to pre-defined sections, while avoiding modifications to the review content and preserving the original itemization if available, along with four in-context learning examples (Section \ref{sec:app-review-itemization}). We evaluate along two key dimensions. As we want to preserve review content, we examine the outputs for hallucinations (added content) and omissions (removed content). As sections play a role in item-level measurements (Section \ref{sec:measurements}) we also examine the performance of item-to-section assignment.

For evaluation, we split $r^f$ and $r^i$ into sentences using \texttt{nltk}, and implement a matching algorithm to align sentences $S^f$ and $S^i$ based on similarity via \texttt{all-MiniLM-L6-v2} embeddings using the \texttt{sentence-transformer} library and $0.8$ as threshold \cite{reimers-2019-sentence-bert, minilm}. We calculate precision as $|S^i \cap S^f|/|S^i|$ and recall as $|S^i \cap S^f|/|S^f|$, approximating the number of hallucinated and omitted sentences, respectively. On a random sample of $30$ reviews from diverse campaigns,\footnote{Over $370$ review items, more than triple of a similar comment-level evaluation in \cite{sadallah-etal-2025-good}.} we obtained precision $P = 0.95$ and recall  $R = 0.88$, indicating strong alignment. The results were further validated by two human raters with computer science expertise, finding no cases of hallucination or omission: most mismatches resulted from sentence-splitting errors and omission of the original review headers by design. 

To assess the quality of item-to-section assignments, we conducted a follow-up human evaluation on the same subsample. Given the transformed review, the original review, and a target item in context, raters judged whether each item was assigned to the correct section. The resulting accuracy is $0.96$ (15 items wrong), most of the errors attributable to items that plausibly belonged to multiple categories. Overall, our evaluation suggests that the LLM-based itemization performs reliably. Section \ref{sec:a:itemeval} details the process and provides further examples of itemization errors.

\section{Quality criteria and measurements}
\label{sec:measurements}



\subsection{Overview}

\begin{table}
\centering
\resizebox{\linewidth}{!}{%
\begin{tabular}{l l l}
\hline
Measurement & Domain & Range \\
\hline

\multicolumn{3}{l}{\textbf{Substantiveness}} \\

$\CIRCLE$ ITX: Core items count $\dagger$ & $i \in r^i$ & $[0; \infty)$ \\
$\Circle$ LEN: Review length  & $r^f$       & $[0; \infty)$ \\
\hline

\multicolumn{3}{l}{\textbf{Actionability}} \\

$\CIRCLE$ ACT: Actionability score $\ddagger$ &$i_w \in r^i$ &  $[0; 5]$ \\
$\Circle$ REQ: Number of requests $\dagger$ & $s \in r^f$ & $[0; \infty)$ \\
\hline

\multicolumn{3}{l}{\textbf{Grounding}} \\

$\CIRCLE$ GND: Grounding score $\ddagger$ &  $i_w \in r^i$  & $[0; 5]$ \\
$\Circle$ EXL: Explicit links $\dagger$  &  $r^f$ & $[0; \infty)$ \\
\hline

\end{tabular}
}
\caption{\textbf{Measurements overview}. Domain: $r^f$ - flattened review, $r^i$ - itemized review, $s$ - review sentences, $i$ review items, $i_w$ - review items in \texttt{Weaknesses+} sections. Pooling: $\dagger$ - count, $\ddagger$ - mean. \texttt{LEN} requires no pooling. Type: $\CIRCLE$ - LLM, $\Circle$ - lightweight.}
\label{tab:metrics-overview}
\vspace{-4mm}
\end{table}

Having transformed the reviews into the unified representations $r^f$ and $r^i$, we compute the measurements of review quality. Following our focus on review utility for authors and editors, we measure quality on three high-level criteria. \textbf{Substantiveness} captures the overall amount of information in a review and serves both authors and editors. \textbf{Actionability} reflects the extent to which the review suggests concrete steps for improving the work, measuring review utility for the authors. \textbf{Grounding} measures whether judgments are explicitly tied to the paper content or prior work, which helps authors address the issues and helps editors validate the reviewers' claims. All three criteria are necessary for high-quality review and well-suited for formal modeling, as they can be estimated from review text and permit verification by a third party. We operationalize these criteria through two types of measurements (Table \ref{tab:metrics-overview}). We propose a set of \textit{lightweight} measurements that are easy to deploy, broadly applicable, and straightforward to interpret. In addition, we explore \textit{LLM-based} measurements that enable deeper analysis, but come at a computational cost. We investigate the relationships between the two measurement types in Section \ref{sec:results}.

\subsection{Length (\texttt{LEN}) and Item count (\texttt{ITX})}

Review length (\texttt{LEN}) is a widely used lightweight proxy of review substantiveness \cite{Bharti2022-pj, Maddi2024-fw, thakkar2025llmfeedbackenhancereview}. We calculate review length as a character count in a flattened peer review, divided by $1000$ for ease of reporting. Despite its efficiency, \texttt{LEN} has important drawbacks: it does not capture the amount of \textit{useful} information in a review, fails to distinguish between specific reviewing issues, and can be misled by uselessly elongated reviews, similar to humans \cite{Goldberg2023-rk}. To address this, we implement \texttt{ITX} which, given an itemized review, counts the number of items in the core informational sections of the review: \texttt{Summary}, \texttt{Strengths} and \texttt{Weaknesses+}. This approach aggregates sentences related to similar issues, and excludes supplementary information in the \texttt{Other} section that can over-inflate character counts.

\subsection{Author utility scores (\texttt{ACT} and \texttt{GND})}
Building upon \citet{sadallah-etal-2025-good}, we integrate two LLM-based measurements of review utility to the authors. Actionability score (\texttt{ACT}) assesses if the reviewers' suggestions can be translated into actions. Grounding score (\texttt{GND}) measures whether reviewer comments relate to specific parts of the paper. Each score is calculated based on review items and ranges between 1 (low) and 5 (high), with labels defined in \citet{sadallah-etal-2025-good}. We use the fine-tuned \texttt{Llama-3.1-8B-Instruct} \citep{grattafiori2024llama3herdmodels} model from \citet{sadallah-etal-2025-good}, which achieves an average quadratic-weighted Cohen’s Kappa $\kappa^{2}$ of $0.51$ (\texttt{ACT}) and $0.53$ (\texttt{GND}) with human scores. 

We extend this work in two crucial ways. First, the original approach requires the reviews to be itemized by the authors and discards the content otherwise. Instead, we use review itemization to detect items in $r^i$, allowing us to process reviews not written in bullet-point style. Second, the original approach only analyses the items in the Weakness sections of peer reviews. As we show (Table \ref{tab:form-fields}), not all reviewing forms contain well-defined Weaknesses. We compute the scores for all items in the \texttt{Weaknesses+} sections of the itemized reviews, and use average as the overall score for each of the measurements.

\subsection{Link (\texttt{EXL}) and Request count (\texttt{REQ})}

Direct references to the paper are a key prerequisite for grounding review claims, reflected in reviewing guidelines at NLP conferences \cite{arr2025reviewerguidelines} and beyond \cite{Rooyen1999}. As a lightweight complement to \texttt{GND}, we implement \texttt{EXL}, which counts the number of explicit links from the review to the paper. Expanding upon the lightweight approaches taken in \cite{rnr, dycke-etal-2023-nlpeer}, we developed a range of string matching patterns to detect explicit links of different types, from sections (\textit{“Sec. 1”}, \textit{“Experiments section”}) and figures (\textit{“Fig. 1”}), to pages (\textit{“p.3”}) and lines (\textit{“L105”}), see Table A.\ref{tab:exl-examples}. To ensure the quality of this precision-oriented measure, we performed three evaluation rounds where one annotator manually explored 100 matches per category from 3000 reviews sampled from all venues at random, and refined the patterns. The final matching algorithm is provided in Figure A.\ref{fig:ref-patterns}; we calculate \texttt{EXL} as a total explicit link count in a flattened review. 

A review that simply states facts about the paper is difficult for the authors to act upon. As a lightweight complement to \texttt{ACT}, we compute \texttt{REQ} by calculating the number of explicit request sentences in a review, including both actions (\textit{“Please add more details on data collection”}) and clarification requests (\textit{“How was the model evaluated?”}). To obtain the metric, we use the pragmatic tagging classifier by \citet{dycke-etal-2023-overview} which classifies each review sentence into one of six classes: \texttt{Recap} (Summary), \texttt{Strength}, \texttt{Weakness}, \texttt{Todo}, \texttt{Structure} and \texttt{Other} -- using the fine-tuned RoBERTa model \cite{liu2019roberta} evaluated at a mean F1 macro score of $80.3$ across multiple domains with different review formats. We split the flattened review into sentences using \texttt{sciSpacy} \citep{neumann-etal-2019-scispacy}, and report \texttt{REQ} as the total number of \texttt{Todo} sentences in the model predictions.

\section{Experimental setup}
\label{sec:setup}

 We access ICLR and NeurIPS data directly via OpenReview in compliance with their Terms of Service, excluding reviews for desk rejected and withdrawn papers. The ARR reviews are derived from the official releases by the ARR data collection initiative and aggregated by year.\footnote{\url{https://arr-data.aclweb.org/}} Table A.\ref{tab:review-counts} provides statistics for the full dataset, and Section \ref{sec:app-data-collection} gives further details on data harvesting and pre-processing. For computational reasons and to avoid large venues dominating the statistics, we conduct our analysis on a random \textbf{sample of 1k reviews} per reviewing campaign, while using the full campaign data for ARR-2022 that has <1k reviews.

Each review was flattened and itemized as described in Section \ref{sec:unification}. Flattening was performed locally, while itemization was done via OpenAI API. This was followed by metric calculation as discussed in Section \ref{sec:measurements}. The LLM-based metrics \texttt{ACT} and \texttt{GND}, as well as the lightweight \texttt{REQ}, were calculated on a compute server. \texttt{ITX} uses the review items predicted during itemization and is computed locally along with \texttt{LEN} and \texttt{EXL}. In total, all measurements take around 4 seconds to compute per review, with most time taken by the LLM-based metrics. Section \ref{sec:app-implementation-details} provides further details.

\textbf{\underline{Note on collection policies.}} Among our target venues, only ICLR makes \emph{all} data openly available. NeurIPS releases reviews for all accepted papers, and allows the authors to opt-in rejected papers and reviews. ARR applies multi-step consent \cite{dycke-etal-2022-yes}, where both authors \textit{and} reviewers need to opt-in, and the paper needs to be accepted. As a result, the data of ARR and NeurIPS shows imbalance towards accepted papers (Table A.\ref{tab:review-counts}). This has important implications, as \textbf{both opt-in and conditioning on paper acceptance can favor higher-quality reviews}. Thus, any measurements of review quality on opt-in venues should be seen as an upper bound, and comparison between venues with different policies should be avoided.
 
\section{Results and analysis}
\label{sec:results}

\subsection{Statistics and Normalization}

Based on the sample data for all reviewing campaigns, the reviews range from $30$ to $28$k characters of text, and on average contain $11$ core items, $4$-$5$ requests and $2$-$3$ explicit links to the paper, with review claims scoring $2.5$ on actionability and $3$ on grounding scale on average. For count-based measurements (see Table \ref{tab:metrics-overview}) and \texttt{LEN}, we observe high variation and a long tail due to the lack of an upper bound, resulting in a positively skewed mean. Figure A.\ref{fig:metric-distributions} reports the value distributions for individual measurements, and Table A.\ref{tab:metric-ranges} gives further statistics. Based on the effective ranges of the metrics, we \textbf{normalize} each metric to $[0, 1]$ for ease of reporting and aggregation.

\subsection{Correlation analysis}

\begin{figure}
    \centering
    \includegraphics[height=4.5cm]{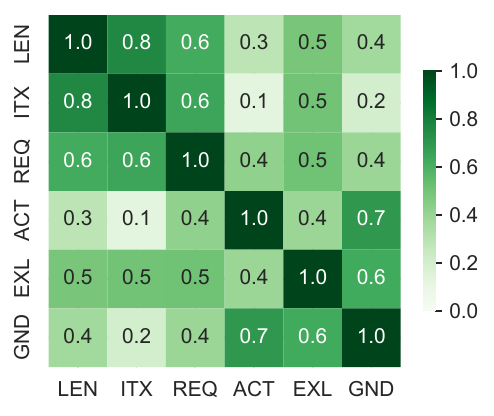} 
    \caption{Metric correlations (Spearman’s $\rho$).}
    \label{fig:metric-correlations}
    \vspace{-2mm}
\end{figure}

Correlation analysis allows us to assess to what extent the metrics reflect related phenomena, and to establish connections between different dimensions of review quality. We find that all measurements are positively correlated (Figure \ref{fig:metric-correlations}), suggesting that our desiderata are non-contradictory and represent a shared construct of review quality. Yet, the correlations are not perfect, suggesting that the measurements are complementary. We note high within-criterion correlations for Substantiveness (\texttt{LEN} to \texttt{ITX}) and Grounding (\texttt{EXL} to \texttt{GND}), and a high correlation between \texttt{ACT} and \texttt{GND} which we partially attribute to both measurements being dependent on the presence of a claim in the review item\ext{QR: ACT and REQ are not well correlated? what are the potential reasons?}.

LLM-based processing is costly and incurs risks due to handling of confidential reviewing data, motivating the search for efficient, locally deployable solutions. Given the positive correlations, we investigate to what extent lightweight measurements can approximate LLM-based measurements in aggregate. We calculate average normalized score per review for each of the two groups (Table \ref{tab:metrics-overview}). We then compare the two resulting aggregate scores in terms of Spearman’s $\rho$, reaching moderately strong $0.6$. This result suggests that while lightweight alternatives do not fully replace LLM-based measurements, they can be deployed inexpensively and at scale to get a first glimpse of reviewing quality.

\subsection{Review quality score}
\label{sec:results-q}

Given that none of the metric pairs show extremely high correlations ($\rho$ > 0.9), we define an aggregate review quality score $Q$ as a non-weighted average of all our measurements (after normalization). To investigate the behavior of the metric, we have manually examined the highest and lowest scoring 1\% of the reviews in the data sample. We found that high-scoring entries originate from diverse campaigns and represent thorough, well-structured reviews with numerous pointers to the paper and actionable suggestions. The composition of the lowest-scoring reviews is mixed: along with less informative and actionable reviews, we find reviews filled with placeholder values, refusals to review due to lack of expertise, and flags for desk rejection -- yet the majority of the low-scoring entries represent genuine reviews.\footnote{The top review with the score of $Q = 0.91$ originates from ICLR-2021 (pre-LLM) and spans \textit{ten pages} of text 12pt single-spaced. In the response, the authors offer the reviewer co-authorship on the paper. The lowest-scoring review comes from ICLR-2019, and has the score of $Q = 0.0$. The review gives the paper a marginally-accepting rating, and its text reads “\textit{I will add it tomorrow}”. No further communication follows. The authors thank the reviewer.}
Figure A.\ref{fig:Q-distribution} reports the distribution of $Q$, skewed due to positive outliers in count-based measurements, and featuring a small set of extremely low-scoring reviews. 

\begin{figure*}[t]
    \centering
    \includegraphics[width=0.85\linewidth]{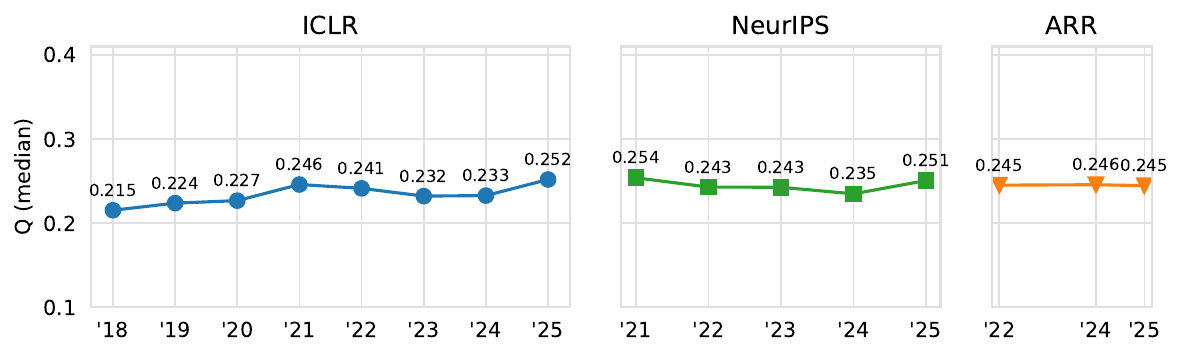}
    \caption{Median aggregate review quality $Q$ across venues and time.}
    \label{fig:Q-venues-time}
\end{figure*}

\subsection{Review quality across venues and time}
\label{sec:rqvt}

As $Q$ distribution is skewed, we compare review campaigns in terms of their \textit{median} $Q$ on our study sample. Figure \ref{fig:Q-venues-time} presents the results. Although we observe some fluctuations in review quality, we see no consistent and substantial decline across years and venues. To test for significance of year-to-year differences, we a apply bootstrap test (10k iterations, 99\% Confidence Interval (CI), Bonferroni correction applied per venue), and find that many of the observed differences are not statistically significant (Section \ref{sec:app-stat-sign}). A few notable exceptions are the \emph{increase} in review quality at ICLR-2021, ICLR-2025 and NeurIPS-2025 compared to preceding years, but the lower limits of the CIs suggest that these differences may be practically negligible. Individual metrics display a similar trend (Section \ref{sec:app-stat-indi}). 
To investigate the effect of selection bias on $Q$, we compare median $Q$ values between ICLR accepted ($0.24$) and rejected papers ($0.23$), indicating a minor decrease (Figure A.\ref{fig:Q-ICLR-acc-vs-rej}). Due to the unavailability of opt-out data, the effects of opt-in policies on $Q$ remain unknown.

\subsection{Validation against human ratings}
\label{sec:humeval}


Finally, we investigate how our constructed, automatic measure of review quality aligns with human judgments. As author-elicited review quality ratings are prone to bias \cite{Goldberg2023-rk}, we require the reference scores to be produced by a third party. We obtained such scores in two complementary, independent ways. From a concurrently released data by \citet{rotrev}, we sampled 100 random reviews from NeurIPS-2023 and ICLR-2024, annotated by human judges with 12 diverse review aspects and an overall quality score. Yet, the inter-annotator agreement on this data is unknown, and our attempts at re-using the provided guidelines resulted in low agreement (see \ref{subsec:app-humeval}). To further validate the result, we developed a new annotation schema and guideline focusing on review value for authors and editors along six numerical ratings averaged into an overall score (see  \ref{sec:app-guidelines}). We then re-annotated the same subset using  the new schema and ensuring multiple annotators per item to calculate agreement (two PhD-level, one faculty-level NLP). On the original annotations from \cite{rotrev} subset, we observe a moderate-strong $0.6$ Spearman correlation between overall quality score and $Q$. On our re-annotated subset that represents \textit{different} annotators using \textit{different} guidelines, we observe a similar moderate to strong correlation ranging from $0.5$ to $0.8$. These results support the credibility of our measurements, and confirm that the framework can serve as a useful approximation of perceived review quality by human raters, while leaving room for improvement.


\section{Discussion}
\label{sec:discussion}

The lack of consistent decline in review quality contradicts the common narrative and invites further discussion. We structure it along five hypotheses: 

\noindent \textbf{H1: Review quality didn't decline.} It is possible that new forms, guidelines and incentives compensated for the negative trends. It is also possible that the reviewing quality would not decline \textit{anyway}. One way to establish the effects of interventions on review quality is through controlled experiments \cite{Goldberg2023-rk, thakkar2025llmfeedbackenhancereview}. Our framework can quantify variation in the outcomes of these experiments, and help find causal links between support measures and review quality.

\noindent \textbf{H2: Review quality \textit{used} to be better.} It is possible that review quality at NeurIPS and ARR was higher prior to 2021. This data is not public, but it exists: for example, SoftConf\footnote{\url{https://softconf.com/}} contains years of earlier data from ACL which could be examined by a trusted party. Our lightweight measurements can be used to process this vast historical data, and review unification ensures that same measurements can be applied to past and future review campaigns.

\noindent \textbf{H3: Our measurements are incomplete.} Quality might have declined along a dimension that we do not measure. While Section~\ref{sec:humeval} indicates moderate-strong correlation to human judgments, there remains space for improvement.
Promising directions include studying the relationships between reviews and papers, alignment between review texts and scores, reviewer's reasoning, and feasibility of the requests. Our study and data can serve as a starting point for the future research on holistic review quality measurement.

\noindent \textbf{H4: Worst reviews got worse.} It is possible that overall quality didn't change, but worst reviews got worse. To investigate, in a supplementary analysis (Section \ref{sec:app-worst-reviews}), we compare median $Q$ of 10\% lowest- and 10\% highest-scoring reviews. While there is fluctuation across both subsets, we see no consistent decline. Yet, this finding is dependent on (H3), and we encourage follow-up work to perform this analysis using new sets of measurements.

\noindent \textbf{H5: More reviews -- more bad reviews.} Finally, the perception of declining review quality might have a statistical explanation. Assuming that the distribution of true review quality $Q^*$ remains constant and the number of reviews grows, the number of \emph{individuals} receiving low-quality reviews will grow proportionally, leading to wasted effort, frustration, rejection of solid work and publication of flawed research. 
In turn, the use of social media and public disclosure of review experiences may intensify negative feelings among rejected authors and negatively affect public perception. Thus, as our communities grow, we need to not only keep reviewing quality as is, but strive to make it better.


\section{Conclusion}

This paper introduced the first holistic, automated framework for the study of review quality, and presented the first study of its kind examining the evolution of review quality across computer science venues and over time. Contrary to the prevailing narrative, our proxy measurements indicate no decline in median review quality at computer science conferences over the recent years. Our follow-up hypotheses outline the rich, open problem space for the future study of review quality.



\section*{Limitations}
\label{sec:limitations}

Our study revealed important limitations of review quality measurement, indicating ample space for future work. While our human evaluation results are encouraging, the \textbf{optimal way to construct and aggregate measurements of review quality} remains an open problem. We experimented with count- and mean-based pooling, and use equal weighting for $Q$ calculation. Alternative aggregation strategies are possible. Given that authors and editors have different -- and at times conflicting -- priorities, future work could explore approaches that prioritize certain dimensions of review quality, or estimate the weights of individual metrics based on extrinsic or survey-based indicators of review quality, such as human ratings or future citations. Our guidelines and data can serve as a starting point for creating larger and more diverse datasets of peer reviews annotated along different dimensions of review quality.

The \textbf{prediction accuracy for the individual metrics} is not perfect, especially for more intricate measurements like the actionability score and the grounding score. As new approaches to these problems emerge, they can be easily integrated into the framework. We note that considering \emph{multiple} measurements of review quality has the potential to mitigate the effects of errors in individual measurements. On a framework scale, the set of quality criteria proposed here might not capture the full spectrum of review issues either, and can be further extended (see Section \ref{sec:discussion}).

Our preliminary results in \textbf{review standardization} indicate that \textit{some} LLMs are capable of unifying peer review structure while preserving the original review content. While the evaluation protocol that we presented in Section \ref{sec:unification} is highly efficient, the absence of gold standard data prevents large-scale formal evaluation and iterative improvement. We deem it plausible that an alternative lightweight approach could be used to perform review itemization at low cost, and leave this exploration to the future: our released data can serve as a silver standard for the evaluation of such approaches.

As we discuss in Section \ref{sec:setup}, due to data publication policies, the analysis of review quality based on open data sources is prone to \textbf{selection bias}. Our preliminary analysis in Section~\ref{sec:rqvt} suggests that acceptance might not have a major effect on review quality, yet it is only based on a single venue that makes such data available -- ICLR -- and needs to be further validated. As opt-out data is not available, the effects of opt-in on review quality are unknown. Investigating these effects requires access to full reviewing data along with opt-in decisions and can be performed by the conference organizers or as part of a community-backed effort, given an appropriate licensing agreement and consent by the reviewers and authors \cite{dycke-etal-2022-yes}.

Our study is \textbf{limited to English} as the main language of international academic publishing. NLP for peer review beyond English is an exciting and under-explored area due to the lack of data: in fact, as of 2026, we are not aware of any major structured datasets in NLP for non-English peer review. We note that the proposed framework does not depend on language and can be adjusted to new languages as the necessary datasets emerge. Similarly, our study \textbf{focuses on computer science conferences} in machine learning and AI. This is due to the availability of data and the domain expertise required to evaluate the results, and does not imply the domain-specificity of the framework. The measurements for \texttt{LEN}, \texttt{EXL}, and \texttt{REQ} can be trivially adopted across domains. Assuming that a peer review in any domain can be converted into a bullet point format, \texttt{ITX} applies across domains as well -- yet, the performance is unknown and would need to be validated. Alternatively, focusing on a single venue (such as ACL) would allow using its specific features (such as Findings vs Main distinction) for a more fine-grained analysis. This is left to the future. Finally, while conceptually \texttt{ACT} and \texttt{GND} apply across domains, the specific measurement tools would require further adaptation, as they depend on models pre-trained in the computer science domain. Note that in any case -- even when applied to English computer science venues -- the normalized metrics should be calibrated given the minimal and maximal values in the data at hand.

Finally, we highlight that the purpose of our framework is to \emph{measure} review quality: it can detect a change, but it can not explain \textit{why} this change occurs. A further step would be to study the \textbf{causes of variation} in review quality (or lack thereof), such as use of AI assistance, reviewers' competence, or the quality of rebuttals and area chair interventions. We note that many such factors require careful analysis and modeling, e.g. AI assistance can come in many varieties, and quantifying reviewer competence or measuring rebuttal quality is by itself a difficult task. Yet, establishing such factors is what ultimately matters if we want to understand and improve our peer review. We thus look forward to future applications of our framework for experimental, causal study of review quality.

\section*{Ethical Considerations}
\label{sec:ethics}

We stress that this is an exploratory study, and our measurement framework is \textbf{NOT} intended for direct practical applications such as low-quality review detection. Such applications carry risks and would require further validation in the community of interest and obligatory human oversight. A particular risk is that, like any other metric, $Q$ and its individual components can be gamed, which should be taken into account when designing reviewing assistance tools. Similarly, our results should \textbf{NOT} be used for policy design without further third-party validation -- yet, they indicate that the question of review quality is worth investigating empirically, and that the popular claim about the declining review quality needs further empirical support. Finally, as discussed in Section \ref{sec:setup}, the differences in data collection policies -- but also in paper complexity, time available to review, reviewing load, and differences in the community culture -- prevent cross-community comparison of review quality, and our results should \textbf{NOT} be read as such.

We note that some research questions about review quality can only be answered by the parties with access to full reviewing data and ability to elicit additional information from peer review stakeholders, such as helpfulness scores and feedback. This creates barriers for researchers with no such access. We encourage policymakers and conference organizers to implement such measures in their communities and make this data available to support research on review quality measurement, e.g., through publication of aggregate statistics or special data access agreements.

All the data used in our study is open and publicly available for research. ARR data is published under a CC-BY license and collected with informed consent from all participating parties. ICLR and NeurIPS data are made available by OpenReview, which applies a CC-BY license on the review reports, effective retroactively.\footnote{\url{https://openreview.net/legal/terms}} Human evaluation of review itemization and review quality was performed by the paper authors and an additional annotator, who consented to the release of their labels and were compensated for the work as part of their regular employment. The annotation tasks carried no risks to the participants.

We have used openly available models whenever possible, and, unlike prior work, explored lightweight approaches to review quality estimation and compared them to LLM-based approaches. Review itemization is the only process in our framework done by a commercial GPT-5.2 model due to its ability to handle long inputs and prompts, and its ability to produce consistent structured output -- our attempts to use an open model (GPT-OSS-120B) have not been successful. Yet, our approach does not rely on any model-specific features of GPT-5.2: as the capabilities of open LLMs improve, review itemization can be transferred to an open model. Moreover, given that the quality of machine itemization is promising (Section \ref{sec:unification}), future studies can use our data as silver data to develop lightweight, open itemization models, potentially even not LLM-based.

During this work, we have used generative AI for minor programming assistance. No AI was used to set research questions, design the experiments, interpret the results, or prepare the publication text.

\section*{Acknowledgments}
This work has been co-funded by the German Research Foundation (DFG) as part of the PEER project (grant GU 798/28-1), European Union (ERC, InterText, 101054961), the German Federal Ministry of Research, Technology and Space, the Hessian Ministry of Higher Education, Research, Science, and the Arts within their joint support of the National Research Center for Applied Cybersecurity ATHENE. Views and opinions expressed are however those of the author(s) only and do not necessarily reflect those of the European Union or the European Research Council. Neither the European Union nor the granting authority can be held responsible for them. We thank Nils Dycke, Qian Ruan and Sheng Lu for their insightful comments and feedback on an earlier draft of this paper.

\bibliography{custom}

\appendix

\clearpage
\section{Appendix}

\subsection{Data collection} \label{sec:app-data-collection}
We crawled the publicly available data for conferences of ICLR (2018 - 2025) and NeurIPS (2021 - 2025) using OpenReview API v1 and v2.\footnote{Accessed Nov 12, 2025.} ARR subsets were aggregated from the official releases of the ARR data collection initiative \citep{dycke-etal-2023-nlpeer, nlpeerARR24, nlpeerARR25}. Since ARR data releases are coupled with conference acceptance, the data is released along major NLP conference acceptance decisions. For ease of reporting, we merge ARR data based on the date the review was written: ARR-2022 includes data from papers accepted to ACL-2022 and NAACL-2022, ARR-2024 covers EMNLP-2024 and NAACL-2025, while ARR-2025 features reviews from papers accepted to ACL-2025. From all campaigns, we omit Desk rejected papers since they often do not contain peer reviews, and Withdrawn papers, since papers are withdrawn for various reasons and the status of the associated materials is unclear. Due to donation-based data collection, ARR data can contain papers with no reviews, which are discarded. Many venues feature different categories of paper acceptance (e.g. “Accept (oral)”, “Accept (poster)”). For simplicity, we consider all of these papers “Accepted” in addition to “Invite to Workshop Track”, and the rest as “Rejected”. Table \ref{tab:review-counts} summarizes the resulting data.

\begin{table}
\centering
\begin{tabular}{lrrr}
\toprule
 & Accept & Reject & Total \\
\midrule
ICLR-2018 & 1.3k & 1.5k & 2.8k \\
ICLR-2019 & 1.5k & 2.8k & 4.3k \\
ICLR-2020 & 2.1k & 4.6k & 6.7k \\
ICLR-2021 & 3.3k & 6.7k & 10.0k \\
ICLR-2022 & 4.2k & 6.0k & 10.2k \\
ICLR-2023 & 6.0k & 8.4k & 14.3k \\
ICLR-2024 & 8.7k & 13.3k & 22.0k \\
ICLR-2025 & 14.9k & 19.9k & 34.9k \\
NeurIPS-2021 & 10.2k & 0.5k & 10.7k \\
NeurIPS-2022 & 9.8k & 0.6k & 10.3k \\
NeurIPS-2023 & 14.4k & 0.8k & 15.2k \\
NeurIPS-2024 & 15.9k & 0.8k & 16.6k \\
NeurIPS-2025 & 21.3k & 1.0k & 22.3k \\
ARR-2022 & 0.7k & - & 0.7k \\
ARR-2024 & 1.9k & - & 1.9k \\
ARR-2025 & 2.4k & - & 2.4k \\
\bottomrule
\end{tabular}

\caption{Number of reviews per campaign, full data.}
\label{tab:review-counts}
\end{table}

\begin{table*}[h]
\centering
\resizebox{.95\textwidth}{!}{%
\begin{tabular}{l p{5cm} p{7cm} p{4cm} p{2cm}}
\toprule
Campaign & Textual & Ordinal & Nominal & Cardinal \\
\midrule
ICLR-2018 & Review & Rating, Confidence & - & - \\
ICLR-2019 & Review & Rating, Confidence & - & - \\
ICLR-2020 & Review & Rating, Experience Assessment, Review Assessment: Thoroughness In Paper Reading, Review Assessment: Checking Correctness Of Derivations And Theory, Review Assessment: Checking Correctness Of Experiments & - & - \\
ICLR-2021 & Review & Rating, Confidence & - & - \\
ICLR-2022 & Summary Of The Paper, Main Review, Summary Of The Review & Recommendation, Confidence, Correctness, Technical Novelty and Significance, Empirical Novelty and Significance & Flag For Ethics Review & - \\
ICLR-2023 & Summary Of The Paper, Strength And Weaknesses, Clarity, Quality, Novelty and Reproducibility, Summary Of The Review & Recommendation, Confidence, Correctness, Technical Novelty And Significance, Empirical Novelty And Significance & Flag For Ethics Review & - \\
ICLR-2024 & Summary, Strengths, Weaknesses, Questions & Rating, Confidence, Presentation, Contribution, Soundness & Flag For Ethics Review, Code Of Conduct & - \\
ICLR-2025 & Summary, Strengths, Weaknesses, Questions & Rating, Confidence, Presentation, Contribution, Soundness & Flag For Ethics Review, Code Of Conduct & - \\
NeurIPS-2021 & Summary, Main Review, Limitations And Societal Impact & Rating, Confidence & Needs Ethics Review, Ethical Concerns, Ethics Review Area, Code Of Conduct & Time Spent Reviewing \\
NeurIPS-2022 & Summary, Strength And Weaknesses, Questions, Limitations & Rating, Confidence, Presentation, Contribution, Soundness & Ethics Flag, Ethics Review Area, Code Of Conduct & - \\
NeurIPS-2023 & Summary, Strengths, Weaknesses, Questions, Limitations & Rating, Confidence, Presentation, Contribution, Soundness & Flag For Ethics Review, Code Of Conduct & - \\
NeurIPS-2024 & Summary, Strengths, Weaknesses, Questions, Limitations & Rating, Confidence, Presentation, Contribution, Soundness & Flag For Ethics Review, Code Of Conduct & - \\
NeurIPS-2025 & Summary, Strength and Weaknesses, Questions, Limitations, Paper Formatting Concerns, Final Justification & Rating, Confidence, Quality, Clarity, Significance, Originality & Ethical Concerns, Code Of Conduct Acknowledgement, Responsible Reviewing Acknowledgement & - \\
ARR-2022 & Paper Summary, Summary Of Strengths, Summary Of Weaknesses, Comments, Suggestions And Typos, Ethical Concerns & Overall, Confidence, Datasets, Software, Author Identity Guess & Best Paper & - \\
ARR-2024 & Paper Summary, Summary Of Strengths, Summary Of Weaknesses, Comments, Suggestions And Typos, Ethical Concerns & Overall Assessment, Confidence, Soundness, Datasets, Software & Best Paper, Knowledge Of Or Educated Guess At Author Identity & - \\
ARR-2025 & Paper Summary, Summary Of Strengths, Summary Of Weaknesses, Comments, Suggestions And Typos, Ethical Concerns & Overall Assessment, Confidence, Soundness, Datasets, Software & Knowledge Of Or Educated Guess At Author Identity & - \\
\bottomrule
\end{tabular}%
}
\caption{Fields in review form across all of the campaigns.}
\label{tab:form-fields}
\end{table*}

\clearpage
\subsection{Review itemization} \label{sec:app-review-itemization}
Review itemization was performed with GPT-5.2 with \textit{reasoning\_effort = medium} and \textit{verbosity = low}. The model response is controlled by passing the JSON schema of a Pydantic object, which contains a list of strings for each of the four sections, with each string representing a single item. Itemizing all reviews cost approximately \$$88$ for the whole study sample (approximately 16k reviews) using OpenAI's batch API on Dec 12, 2025. The prompt is given in Section \ref{sec:app-prompt}

\subsection{Implementation details} \label{sec:app-implementation-details}
For the LLM-based author utility metrics \texttt{ACT} and \texttt{GND}, inference took approximately 15 hours on one A100 GPU; \texttt{ITX} directly re-uses the review items predicted during itemization and does not incur further costs. Among the lightweight metrics, \texttt{REQ} was calculated using a RoBERTa model, which took around 1 hour on an A100 GPU, but can be deployed on a CPU machine. \texttt{LEN} and \texttt{EXL} were calculated locally on a CPU-only laptop within seconds. Generating all scores for a single review takes approximately 4 seconds.

\begin{table}[!h]
\centering
\begin{tabular}{lp{0.6\linewidth}}
\hline
\textbf{Category} & \textbf{Examples} \\
\hline
Section    &  \textit{sec.5}, \textit{experiments section} \\
Table      &  \textit{Table 1}, \textit{Tab.1}\\
Figure     &  \textit{Figure 2}, \textit{Fig. 2}, \textit{Fig. 2a}, \\
Equation   &  \textit{Equation 5}, \textit{Eq.5}, \textit{Eq. (5)}\\
Algorithm  &  \textit{Algorithm 1}, \textit{Alg. 1}\\
Line       &  \textit{Line 100}, \textit{L545-547}, \textit{l. 66}, \\
Page       &  \textit{Page 3}, \textit{P.3}\\
\hline
\end{tabular}
\caption{Explicit link examples by category. Omitted for brevity: paragraph, appendix, theorem, footnote, lemma, formula.}
\label{tab:exl-examples}
\end{table}

\begin{figure*}[h]
    \centering
    \begin{Verbatim}[frame=single, fontsize=\small]
section: \b(?:sections?|sec\.?|§|chapter|subsection)\s*[0-9]+(?:\.[0-9]+)*|
\b(?:introduction|related work|method(?:s|ology)?|approach|experiments?|results?|analysis|
discussion|limitations?|conclusion|ablation|future work)\s+section\b
table: \b(?:table|tab\.)\s*[0-9]+[a-zA-Z]?
figure: \b(?:figure|fig\.)\s*[0-9]+[a-zA-Z]?
theorem: \btheorem\s*[0-9]+[a-zA-Z]?
lemma: \blemma\s*[0-9]+[a-zA-Z]?
equation: \b(?:equation|eq\.)\s*\(?[0-9]+(?:\.[0-9]+)?\)?
formula: \bformula\s*\(?[0-9]+(?:\.[0-9]+)?\)?
algorithm: \b(algorithm|alg\.?)\s*\(?[0-9]+(?:\.[0-9]+)?\)?
paragraph: \bparagraphs?\b
line: (?<![a-zA-Z])(?:line|lines|l\.?|L)\s*[\(\s]*[0-9]+(?:\s*(?:-|-|-|,|to)\s*[0-9]+)*[\)\s]*
page: (?<![a-z])(?:page|p\.)\s*[0-9]+
footnote: \bfootnote\s*[0-9]+
appendix: \bappendix\b
  \end{Verbatim}
    \caption{Explicit linking patterns used to calculate \texttt{EXL}. In the \textit{line} category, L0-L4 are blacklisted since they predominantly denote regularization, language proficiency, math symbols, etc. All patterns are case-insensitive.}
    \label{fig:ref-patterns}
\end{figure*}



\clearpage
\onecolumn
\subsection{Additional results}

\begin{figure*}[!h]
    \centering
    \includegraphics[width=\textwidth]{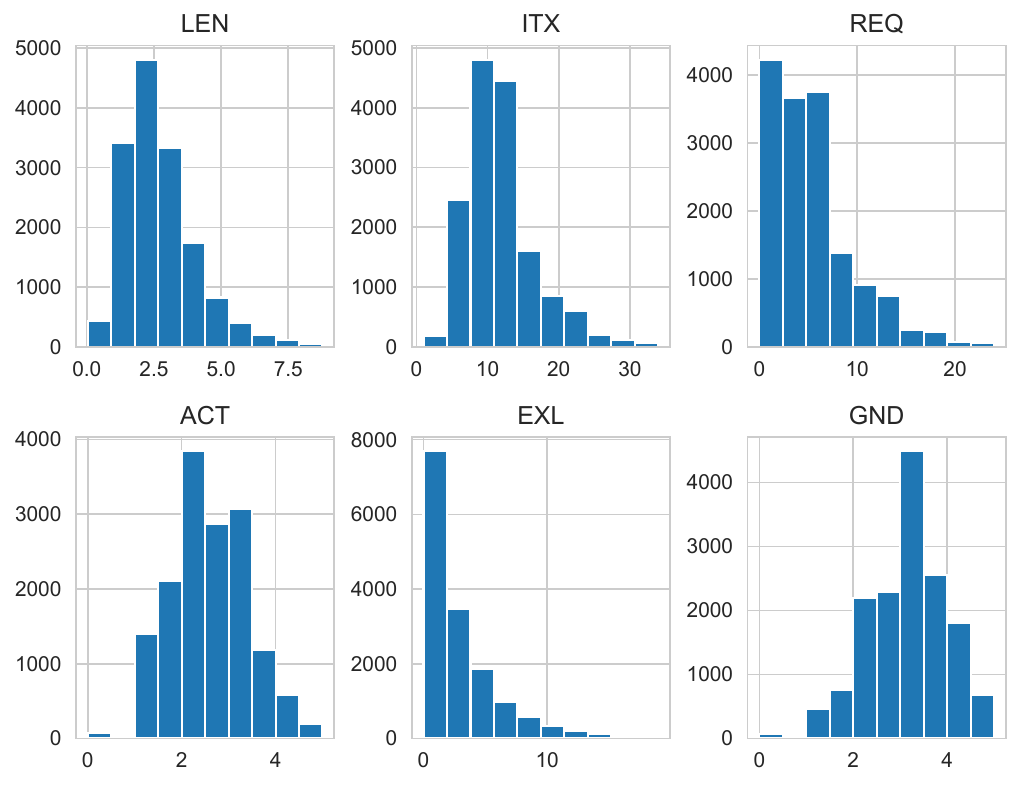}
    \caption{Metric distributions for all the considered measurements. Top-1\% reviews on count-based columns and \texttt{LEN} are excluded for better comprehension (380 out of 15683 reviews). Note the small number of reviews scoring 0 on \texttt{ACT} and \texttt{GND} scores correspond to the cases where the review did not contain a single claim to be scored. All other statistics and results in the paper are reported on the full sample distribution unless specified otherwise.}
    \label{fig:metric-distributions}
\end{figure*}

\twocolumn
\clearpage

\begin{table}
\centering
\begin{tabular}{lrrrr}
\toprule
 & min & max & median & mean \\
\midrule
LEN & 0.03 & 28.29 & 2.49 & 2.85 \\
ITX & 1.00 & 99.00 & 11.00 & 12.19 \\
REQ & 0.00 & 72.00 & 4.00 & 5.73 \\
ACT & 0.00 & 5.00 & 2.50 & 2.53 \\
EXL & 0.00 & 103.00 & 2.00 & 2.97 \\
GND & 0.00 & 5.00 & 3.00 & 3.08 \\
\bottomrule
\end{tabular}

\caption{Measurement statistics.}
\label{tab:metric-ranges}
\vspace{-4mm}
\end{table}

\begin{figure}
    \centering
    \includegraphics[height=6cm]{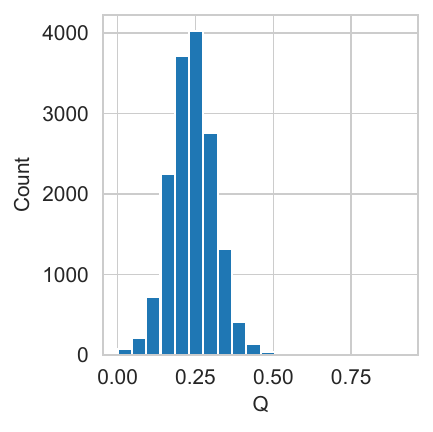}
    \caption{$Q$ distribution on the study sample, all venues.}
    \label{fig:Q-distribution}
\end{figure}

\subsubsection{Statistical testing on overall data} \label{sec:app-stat-sign}
We used \texttt{scikit-learn} library for the bootstrap tests. $H_0$ represents the hypothesis that there is no change in the median $Q$ between any two years. The tables below display the CI for the difference in medians between the year in the column and the row across all the years for each venue. We reject $H_0$ if the CI does not contain 0, and highlight the interval in bold.

\onecolumn
\begin{table*}[htbp]
\centering
\resizebox{.95\textwidth}{!}{%
\begin{tabular}{llllllll}
\toprule
 & ICLR-2019 & ICLR-2020 & ICLR-2021 & ICLR-2022 & ICLR-2023 & ICLR-2024 & ICLR-2025 \\
\midrule
ICLR-2018 & [-0.004, 0.022] & [-0.002, 0.028] & \textbf{[0.017, 0.046]} & \textbf{[0.011, 0.040]} & \textbf{[0.003, 0.031]} & \textbf{[0.005, 0.032]} & \textbf{[0.023, 0.051]} \\
ICLR-2019 & - & [-0.008, 0.015] & \textbf{[0.009, 0.035]} & \textbf{[0.005, 0.032]} & [-0.003, 0.023] & [-0.002, 0.020] & \textbf{ [0.017, 0.040]} \\
ICLR-2020 & - & - & \textbf{[0.006, 0.031]} & \textbf{[0.002, 0.026]} & [-0.007, 0.018] & [-0.007, 0.017] & \textbf{[0.012, 0.036]} \\
ICLR-2021 & - & - & - & [-0.017, 0.010] & \textbf{[-0.025, 0.000]} & \textbf{[-0.026, 0.000]} & [-0.009, 0.019] \\
ICLR-2022 & - & - & - & - & [-0.022, 0.004] & [-0.021, 0.003] & [-0.003, 0.022] \\
ICLR-2023 & - & - & - & - & - & [-0.011, 0.013] & \textbf{[0.006, 0.034]} \\
ICLR-2024 & - & - & - & - & - & - & \textbf{[0.008, 0.030]} \\
\bottomrule
\end{tabular}
}
\caption{CIs for the pairwise difference in medians across the years for ICLR. The interval is highlighted in bold if the difference is significant. Note that an interval being close to 0 indicates no practical difference in the medians.}
\label{tab:iclr-stat}
\end{table*}

\begin{table*}[htbp]
\centering
\resizebox{.65\textwidth}{!}{%
\begin{tabular}{lllll}
\toprule
& NeurIPS-2022 & NeurIPS-2023 & NeurIPS-2024 & NeurIPS-2025 \\
\midrule
NeurIPS-2021 & [-0.022, 0.002] & [-0.025, 0.002] & \textbf{[-0.032, -0.006]} & [-0.015, 0.010] \\
NeurIPS-2022 & - & [-0.013, 0.012] & [-0.021, 0.005] & [-0.003, 0.020] \\
NeurIPS-2023 & - & - & [-0.022, 0.006] & [-0.004, 0.022] \\
NeurIPS-2024 & - & - & - & \textbf{[0.003, 0.029]} \\
\bottomrule
\end{tabular}
}
\caption{CIs for the pairwise difference in medians across the years for NeurIPS. The interval is highlighted in bold if the difference is significant. Note that an interval being close to 0 indicates no practical difference in the medians.}
\label{tab:neurips-stat}
\end{table*}

\begin{table*}[htbp]
\centering
\resizebox{.35\textwidth}{!}{%
\begin{tabular}{lll}
\toprule
 & ARR-2024 & ARR-2025 \\
\midrule
ARR-2022 & [-0.013, 0.017] & [-0.015, 0.015] \\
ARR-2024 & - & [-0.014, 0.010] \\
\bottomrule
\end{tabular}
}
\caption{CIs for the pairwise difference in medians across the years for ARR. The interval is highlighted in bold if the difference is significant. Note that an interval being close to 0 indicates no practical difference in the medians.}
\label{tab:arr-stat}
\end{table*}

\subsubsection{Statistical Testing on individual measurements} \label{sec:app-stat-indi}

\begin{figure*}[!h]
    \centering
    \includegraphics[width=\linewidth]{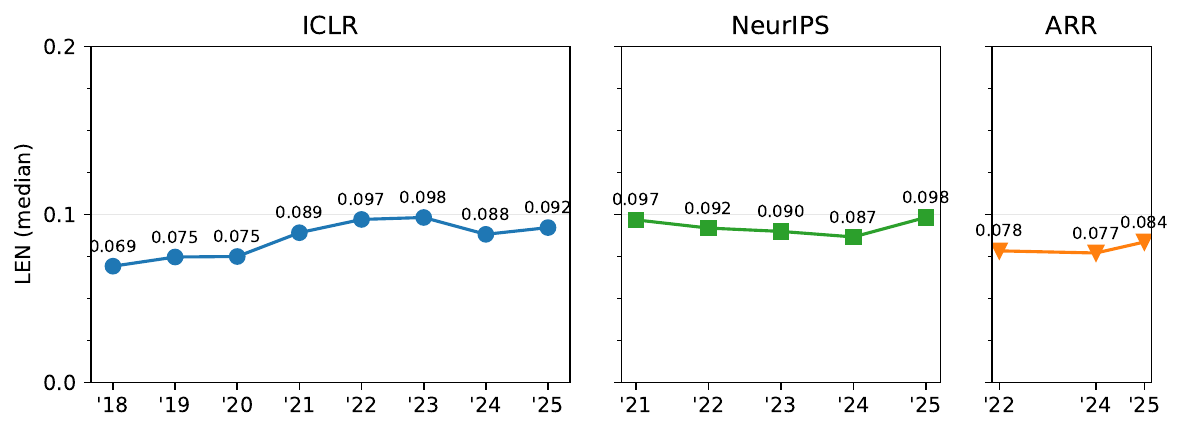}
    \caption{LEN across venues and time.}
    \label{fig:LEN-venues-time}
\end{figure*}

\begin{table}[H]
\centering
\resizebox{.95\textwidth}{!}{%
\begin{tabular}{llllllll}
\toprule
 & ICLR-2019 & ICLR-2020 & ICLR-2021 & ICLR-2022 & ICLR-2023 & ICLR-2024 & ICLR-2025 \\
\midrule
ICLR-2018 & [-0.004, 0.015] & [-0.002, 0.014] & \textbf{[0.011, 0.029]} & \textbf{[0.019, 0.036]} & \textbf{[0.020, 0.039]} & \textbf{[0.009, 0.027]} & \textbf{[0.015, 0.031]} \\
ICLR-2019 & - & [-0.008, 0.009] & \textbf{[0.005, 0.024]} & \textbf{[0.012, 0.031]} & \textbf{[0.014, 0.034]} & \textbf{[0.003, 0.021]} & \textbf{[0.009, 0.026]} \\
ICLR-2020 & - & - & \textbf{[0.004, 0.023]} & \textbf{[0.012, 0.030]} & \textbf{[0.014, 0.034]} & \textbf{[0.003, 0.020]} & \textbf{[0.009, 0.025]} \\
ICLR-2021 & - & - & - & [-0.002, 0.016] & [0.000, 0.019] & [-0.010, 0.007] & [-0.005, 0.011] \\
ICLR-2022 & - & - & - & - & [-0.007, 0.010] & \textbf{[-0.019, 0.000]} & [-0.012, 0.004] \\
ICLR-2023 & - & - & - & - & - & \textbf{[-0.021, -0.002]} & [-0.016, 0.003] \\
ICLR-2024 & - & - & - & - & - & - & [-0.004, 0.014] \\
\bottomrule
\end{tabular}
}
\caption{CIs for the pairwise difference in medians of LEN across the years for ICLR. The interval is highlighted in bold if the difference is significant. Note that an interval being close to 0 indicates no practical difference in the medians.}
\label{tab:iclr-stat-LEN}
\end{table}

\begin{table}[H]
\centering
\resizebox{.65\textwidth}{!}{%
\begin{tabular}{lllll}
\toprule
& NeurIPS-2022 & NeurIPS-2023 & NeurIPS-2024 & NeurIPS-2025 \\
\midrule
NeurIPS-2021 & [-0.013, 0.003] & [-0.015, 0.0] & \textbf{[-0.017, -0.003]} & [-0.007, 0.010] \\
NeurIPS-2022 & - & [-0.01, 0.005] & [-0.012, 0.002] & [-0.001, 0.014] \\
NeurIPS-2023 & - & - & [-0.009, 0.003] & \textbf{[0.001, 0.015]} \\
NeurIPS-2024 & - & - & - & \textbf{[0.004, 0.018]} \\
\bottomrule
\end{tabular}
}
\caption{CIs for the pairwise difference in medians of LEN across the years for NeurIPS. The interval is highlighted in bold if the difference is significant. Note that an interval being close to 0 indicates no practical difference in the medians.}
\label{tab:neurips-stat-LEN}
\end{table}

\begin{table}[H]
\centering
\resizebox{.35\textwidth}{!}{%
\begin{tabular}{lll}
\toprule
 & ARR-2024 & ARR-2025 \\
\midrule
ARR-2022 & [-0.009, 0.005] & [-0.003, 0.013] \\
ARR-2024 & - & \textbf{[0.001, 0.013]} \\
\bottomrule
\end{tabular}
}
\caption{CIs for the pairwise difference in medians of LEN across the years for ARR. The interval is highlighted in bold if the difference is significant. Note that an interval being close to 0 indicates no practical difference in the medians.}
\label{tab:arr-stat-LEN}
\end{table}

\begin{figure*}[!h]
    \centering
    \includegraphics[width=\linewidth]{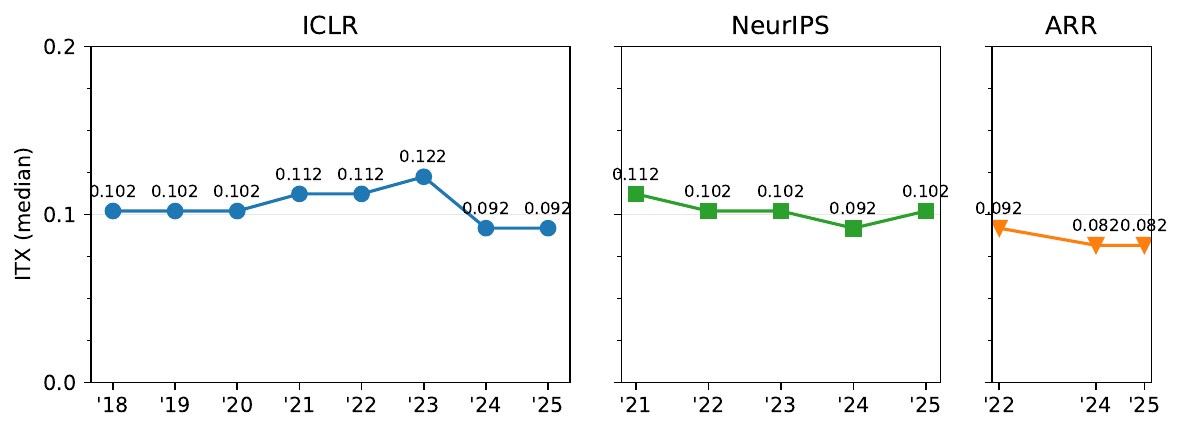}
    \caption{ITX across venues and time.}
    \label{fig:ITX-venues-time}
\end{figure*}

\begin{table}[H]
\centering
\resizebox{.95\textwidth}{!}{%
\begin{tabular}{llllllll}
\toprule
 & ICLR-2019 & ICLR-2020 & ICLR-2021 & ICLR-2022 & ICLR-2023 & ICLR-2024 & ICLR-2025 \\
\midrule
ICLR-2018 & [-0.010, 0.010] & [-0.010, 0.010] & [0.000, 0.020] & [0.000, 0.031] & \textbf{[0.020, 0.041]} & [-0.010, 0.010] & [-0.010, 0.010] \\
ICLR-2019 & - & [-0.010, 0.010] & [0.000, 0.020] & [0.000, 0.031] & \textbf{[0.020, 0.041]} & [-0.010, 0.010] & [-0.010, 0.010] \\
ICLR-2020 & - & - & [0.000, 0.020] & [0.000, 0.031] & \textbf{[0.020, 0.041]} & [-0.010, 0.010] & [-0.010, 0.010] \\
ICLR-2021 & - & - & - & [-0.010, 0.020] & \textbf{[0.010, 0.031]} & [-0.020, 0.000] & [-0.020, 0.000] \\
ICLR-2022 & - & - & - & - & [0.000, 0.023] & \textbf{[-0.031, -0.010]} & [-0.031, 0.000] \\
ICLR-2023 & - & - & - & - & - & \textbf{[-0.041, -0.020]} & \textbf{[-0.041, -0.020]} \\
ICLR-2024 & - & - & - & - & - & - & [-0.010, 0.010] \\
\bottomrule
\end{tabular}
}
\caption{CIs for the pairwise difference in medians of ITX across the years for ICLR. The interval is highlighted in bold if the difference is significant. Note that an interval being close to 0 indicates no practical difference in the medians.}
\label{tab:iclr-stat-ITX}
\end{table}

\begin{table}[H]
\centering
\resizebox{.65\textwidth}{!}{%
\begin{tabular}{lllll}
\toprule
& NeurIPS-2022 & NeurIPS-2023 & NeurIPS-2024 & NeurIPS-2025 \\
\midrule
NeurIPS-2021 & [-0.020, 0.000] & \textbf{[-0.031, -0.010]} & \textbf{[-0.031, -0.010]} & \textbf{[-0.031, -0.010]} \\
NeurIPS-2022 & - & [-0.010, 0.010] & [-0.010, 0.010] & [-0.010, 0.010] \\
NeurIPS-2023 & - & - & [-0.010, 0.010] & [-0.010, 0.010] \\
NeurIPS-2024 & - & - & - & [-0.010, 0.010] \\
\bottomrule
\end{tabular}
}
\caption{CIs for the pairwise difference in medians of ITX across the years for NeurIPS. The interval is highlighted in bold if the difference is significant. Note that an interval being close to 0 indicates no practical difference in the medians.}
\label{tab:neurips-stat-ITX}
\end{table}

\begin{table}[H]
\centering
\resizebox{.35\textwidth}{!}{%
\begin{tabular}{lll}
\toprule
 & ARR-2024 & ARR-2025 \\
\midrule
ARR-2022 & [-0.020, 0.000] & [-0.020, 0.000] \\
ARR-2024 & - & [-0.010, 0.010] \\
\bottomrule
\end{tabular}
}
\caption{CIs for the pairwise difference in medians of ITX across the years for ARR. The interval is highlighted in bold if the difference is significant. Note that an interval being close to 0 indicates no practical difference in the medians.}
\label{tab:arr-stat-ITX}
\end{table}

\begin{figure*}[!h]
    \centering
    \includegraphics[width=\linewidth]{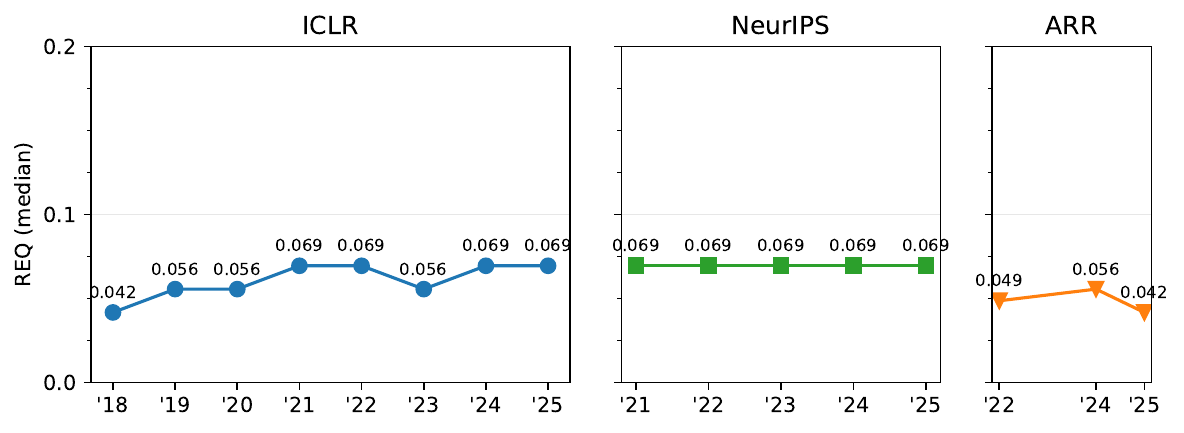}
    \caption{REQ across venues and time.}
    \label{fig:REQ-venues-time}
\end{figure*}

\begin{table}[H]
\centering
\resizebox{.95\textwidth}{!}{%
\begin{tabular}{llllllll}
\toprule
 & ICLR-2019 & ICLR-2020 & ICLR-2021 & ICLR-2022 & ICLR-2023 & ICLR-2024 & ICLR-2025 \\
\midrule
ICLR-2018 & [-0.014, 0.015] & [0.000, 0.028] & [0.000, 0.028] & [0.000, 0.028] & [0.000, 0.028] & [0.000, 0.028] & \textbf{[0.014, 0.042]} \\
ICLR-2019 & - & [-0.008, 0.014] & [-0.001, 0.028] & [0.000, 0.028] & [-0.008, 0.014] & [0.000, 0.028] & \textbf{[0.005, 0.042]} \\
ICLR-2020 & - & - & [-0.014, 0.014] & [0.000, 0.014] & [-0.014, 0.014] & [0.000, 0.014] & [0.000, 0.028] \\
ICLR-2021 & - & - & - & [-0.014, 0.014] & [-0.014, 0.014] & [-0.014, 0.014] & [0.000, 0.028] \\
ICLR-2022 & - & - & - & - & [-0.014, 0.014] & [-0.014, 0.014] & [0.000, 0.028] \\
ICLR-2023 & - & - & - & - & - & [-0.008, 0.014] & [0.000, 0.028] \\
ICLR-2024 & - & - & - & - & - & - & [0.000, 0.028] \\
\bottomrule
\end{tabular}
}
\caption{CIs for the pairwise difference in medians of REQ across the years for ICLR. The interval is highlighted in bold if the difference is significant. Note that an interval being close to 0 indicates no practical difference in the medians.}
\label{tab:iclr-stat-REQ}
\end{table}

\begin{table}[H]
\centering
\resizebox{.65\textwidth}{!}{%
\begin{tabular}{lllll}
\toprule
& NeurIPS-2022 & NeurIPS-2023 & NeurIPS-2024 & NeurIPS-2025 \\
\midrule
NeurIPS-2021 & [0.000, 0.028] & [0.000, 0.028] & [-0.014, 0.014] & [0.000, 0.014] \\
NeurIPS-2022 & - & [-0.014, 0.014] & [-0.014, 0.000] & [-0.014, 0.014] \\
NeurIPS-2023 & - & - & [-0.014, 0.000] & [-0.014, 0.014] \\
NeurIPS-2024 & - & - & - & [0.000, 0.014] \\
\bottomrule
\end{tabular}
}
\caption{CIs for the pairwise difference in medians of REQ across the years for NeurIPS. The interval is highlighted in bold if the difference is significant. Note that an interval being close to 0 indicates no practical difference in the medians.}
\label{tab:neurips-stat-REQ}
\end{table}

\begin{table}[H]
\centering
\resizebox{.35\textwidth}{!}{%
\begin{tabular}{lll}
\toprule
 & ARR-2024 & ARR-2025 \\
\midrule
ARR-2022 & [-0.014, 0.014] & [-0.014, 0.014] \\
ARR-2024 & - & [-0.014, 0.014] \\
\bottomrule
\end{tabular}
}
\caption{CIs for the pairwise difference in medians of REQ across the years for ARR. The interval is highlighted in bold if the difference is significant. Note that an interval being close to 0 indicates no practical difference in the medians.}
\label{tab:arr-stat-REQ}
\end{table}

\begin{figure*}[!h]
    \centering
    \includegraphics[width=\linewidth]{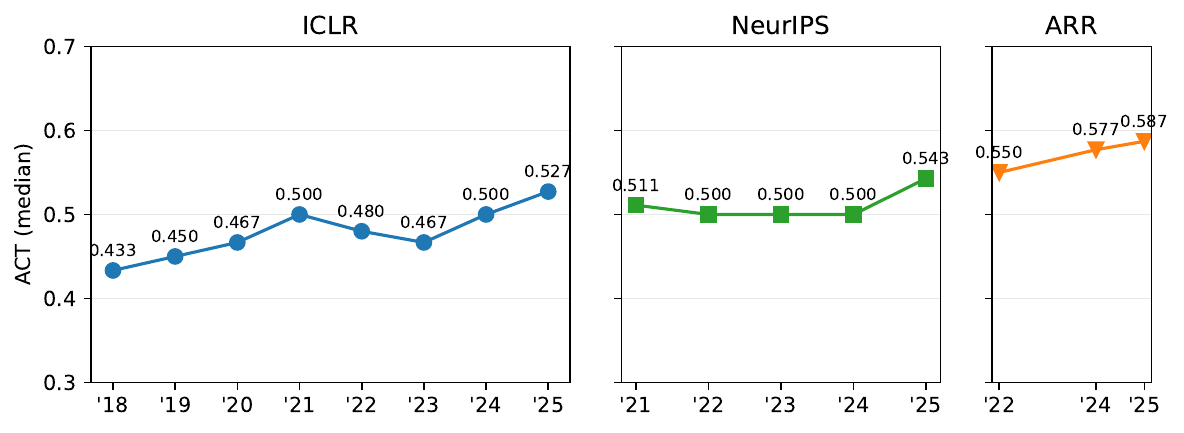}
    \caption{ACT across venues and time.}
    \label{fig:ACT-venues-time}
\end{figure*}

\begin{table}[H]
\centering
\resizebox{.95\textwidth}{!}{%
\begin{tabular}{llllllll}
\toprule
 & ICLR-2019 & ICLR-2020 & ICLR-2021 & ICLR-2022 & ICLR-2023 & ICLR-2024 & ICLR-2025 \\
\midrule
ICLR-2018 & [-0.016, 0.067] & \textbf{[0.004, 0.082]} & \textbf{[0.025, 0.111]} & \textbf{[0.017, 0.095]} & [-0.004, 0.077] & \textbf{[0.027, 0.100]} & \textbf{[0.065, 0.143]} \\
ICLR-2019 & - & [-0.017, 0.051] & \textbf{[0.008, 0.083]} & [0.000, 0.067] & [-0.017, 0.043] & \textbf{[0.007, 0.072]} & \textbf{[0.047, 0.111]} \\
ICLR-2020 & - & - & [-0.003, 0.063] & [-0.020, 0.050] & [-0.032, 0.027] & [-0.006, 0.056] & \textbf{[0.030, 0.093]} \\
ICLR-2021 & - & - & - & [-0.051, 0.024] & [-0.069, 0.003] & [-0.039, 0.033] & [-0.002, 0.070] \\
ICLR-2022 & - & - & - & - & [-0.050, 0.013] & [-0.024, 0.040] & \textbf{[0.019, 0.083]} \\
ICLR-2023 & - & - & - & - & - & [0.000, 0.060] & \textbf{[0.034, 0.100]} \\
ICLR-2024 & - & - & - & - & - & - & \textbf{[0.011, 0.068]} \\
\bottomrule
\end{tabular}
}
\caption{CIs for the pairwise difference in medians of ACT across the years for ICLR. The interval is highlighted in bold if the difference is significant. Note that an interval being close to 0 indicates no practical difference in the medians.}
\label{tab:iclr-stat-ACT}
\end{table}

\begin{table}[H]
\centering
\resizebox{.65\textwidth}{!}{%
\begin{tabular}{lllll}
\toprule
& NeurIPS-2022 & NeurIPS-2023 & NeurIPS-2024 & NeurIPS-2025 \\
\midrule
NeurIPS-2021 & [-0.045, 0.010] & [-0.033, 0.021] & [-0.040, 0.020] & [0.000, 0.061] \\
NeurIPS-2022 & - & [-0.014, 0.040] & [-0.020, 0.040] & \textbf{[0.020, 0.081]} \\
NeurIPS-2023 & - & - & [-0.040, 0.023] & \textbf{[0.005, 0.065]} \\
NeurIPS-2024 & - & - & - & \textbf{[0.004, 0.076]} \\
\bottomrule
\end{tabular}
}
\caption{CIs for the pairwise difference in medians of ACT across the years for NeurIPS. The interval is highlighted in bold if the difference is significant. Note that an interval being close to 0 indicates no practical difference in the medians.}
\label{tab:neurips-stat-ACT}
\end{table}

\begin{table}[H]
\centering
\resizebox{.35\textwidth}{!}{%
\begin{tabular}{lll}
\toprule
 & ARR-2024 & ARR-2025 \\
\midrule
ARR-2022 & [-0.031, 0.080] & [-0.022, 0.080] \\
ARR-2024 & - & [-0.050, 0.057] \\
\bottomrule
\end{tabular}
}
\caption{CIs for the pairwise difference in medians of ACT across the years for ARR. The interval is highlighted in bold if the difference is significant. Note that an interval being close to 0 indicates no practical difference in the medians.}
\label{tab:arr-stat-ACT}
\end{table}

\begin{figure*}[!h]
    \centering
    \includegraphics[width=\linewidth]{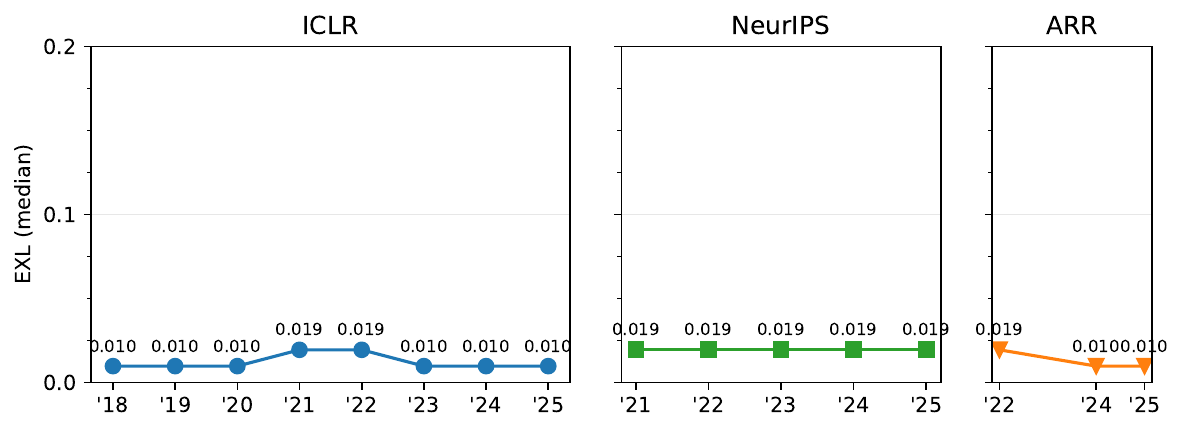}
    \caption{EXL across venues and time.}
    \label{fig:EXL-venues-time}
\end{figure*}

\begin{table}[H]
\centering
\resizebox{.95\textwidth}{!}{%
\begin{tabular}{llllllll}
\toprule
 & ICLR-2019 & ICLR-2020 & ICLR-2021 & ICLR-2022 & ICLR-2023 & ICLR-2024 & ICLR-2025 \\
\midrule
ICLR-2018 & [-0.001, 0.010] & [-0.001, 0.010] & \textbf{[0.005, 0.010]} & \textbf{[0.009, 0.010]} & [-0.001, 0.010] & [-0.001, 0.010] & [-0.001, 0.010] \\
ICLR-2019 & - & [-0.010, 0.010] & [0.000, 0.010] & [0.000, 0.010] & [-0.010, 0.010] & [-0.010, 0.010] & [-0.010, 0.010] \\
ICLR-2020 & - & - & [0.000, 0.010] & [0.000, 0.010] & [-0.010, 0.010] & [-0.010, 0.010] & [-0.010, 0.010] \\
ICLR-2021 & - & - & - & [0.000, 0.000] & [-0.010, 0.000] & [-0.010, 0.000] & [-0.010, 0.000] \\
ICLR-2022 & - & - & - & - & [-0.010, 0.000] & [-0.010, 0.000] & [-0.010, 0.000] \\
ICLR-2023 & - & - & - & - & - & [-0.010, 0.010] & [-0.010, 0.010] \\
ICLR-2024 & - & - & - & - & - & - & [-0.010, 0.010] \\
\bottomrule
\end{tabular}
}
\caption{CIs for the pairwise difference in medians of EXL across the years for ICLR. The interval is highlighted in bold if the difference is significant. Note that an interval being close to 0 indicates no practical difference in the medians.}
\label{tab:iclr-stat-EXL}
\end{table}

\begin{table}[H]
\centering
\resizebox{.65\textwidth}{!}{%
\begin{tabular}{lllll}
\toprule
& NeurIPS-2022 & NeurIPS-2023 & NeurIPS-2024 & NeurIPS-2025 \\
\midrule
NeurIPS-2021 & [-0.010, 0.000] & [-0.019, 0.000] & [-0.019, 0.000] & [-0.019, 0.000] \\
NeurIPS-2022 & - & [-0.010, 0.000] & [-0.010, 0.00] & [-0.010, 0.000] \\
NeurIPS-2023 & - & - & [-0.010, 0.010] & [-0.010, 0.010] \\
NeurIPS-2024 & - & - & - & [-0.010, 0.010] \\
\bottomrule
\end{tabular}
}
\caption{CIs for the pairwise difference in medians of EXL across the years for NeurIPS. The interval is highlighted in bold if the difference is significant. Note that an interval being close to 0 indicates no practical difference in the medians.}
\label{tab:neurips-stat-EXL}
\end{table}

\begin{table}[H]
\centering
\resizebox{.35\textwidth}{!}{%
\begin{tabular}{lll}
\toprule
 & ARR-2024 & ARR-2025 \\
\midrule
ARR-2022 & [-0.010, 0.000] & [-0.010, 0.000] \\
ARR-2024 & - & [-0.010, 0.000] \\
\bottomrule
\end{tabular}
}
\caption{CIs for the pairwise difference in medians of EXL across the years for ARR. The interval is highlighted in bold if the difference is significant. Note that an interval being close to 0 indicates no practical difference in the medians.}
\label{tab:arr-stat-EXL}
\end{table}

\begin{figure*}[!h]
    \centering
    \includegraphics[width=\linewidth]{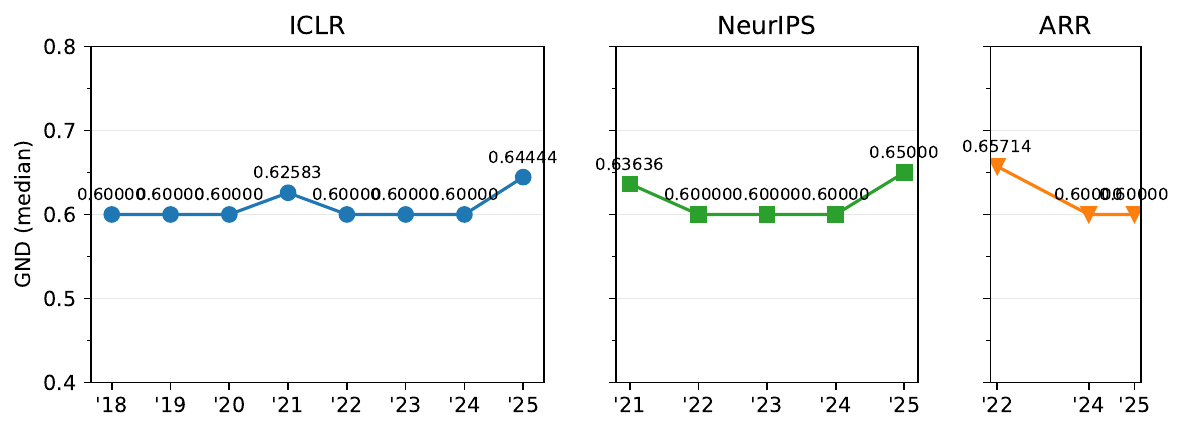}
    \caption{GND across venues and time.}
    \label{fig:GND-venues-time}
\end{figure*}

\begin{table}[H]
\centering
\resizebox{.95\textwidth}{!}{%
\begin{tabular}{llllllll}
\toprule
 & ICLR-2019 & ICLR-2020 & ICLR-2021 & ICLR-2022 & ICLR-2023 & ICLR-2024 & ICLR-2025 \\
\midrule
ICLR-2018 & [-0.036, 0.040] & [0.000, 0.040] & [0.000, 0.076] & [0.000, 0.044] & [-0.040, 0.040] & [0.000, 0.040] & [0.000, 0.094] \\
ICLR-2019 & - & [0.000, 0.036] & [0.000, 0.068] & [0.000, 0.040] & [-0.040, 0.036] & [0.000, 0.036] & [0.000, 0.080] \\
ICLR-2020 & - & - & [0.000, 0.052] & [0.000, 0.039] & [-0.040, 0.000] & [0.000, 0.000] & [0.000, 0.067] \\
ICLR-2021 & - & - & - & [-0.050, 0.032] & [-0.073, 0.000] & [-0.050, 0.000] & [-0.047, 0.067] \\
ICLR-2022 & - & - & - & - & [-0.040, 0.000] & [-0.037, 0.000] & [-0.024, 0.067] \\
ICLR-2023 & - & - & - & - & - & [0.00, 0.040] & [0.000, 0.092] \\
ICLR-2024 & - & - & - & - & - & - & [0.000, 0.067] \\
\bottomrule
\end{tabular}
}
\caption{CIs for the pairwise difference in medians of GND across the years for ICLR. The interval is highlighted in bold if the difference is significant. Note that an interval being close to 0 indicates no practical difference in the medians.}
\label{tab:iclr-stat-GND}
\end{table}

\begin{table}[H]
\centering
\resizebox{.65\textwidth}{!}{%
\begin{tabular}{lllll}
\toprule
& NeurIPS-2022 & NeurIPS-2023 & NeurIPS-2024 & NeurIPS-2025 \\
\midrule
NeurIPS-2021 & [-0.067, 0.044] & [-0.067, 0.040] & [-0.067, 0.000] & [-0.044, 0.067] \\
NeurIPS-2022 & - & [-0.050, 0.044] & [-0.050, 0.000] & [-0.030, 0.067] \\
NeurIPS-2023 & - & - & [-0.050, 0.000] & [-0.010, 0.067] \\
NeurIPS-2024 & - & - & - & [0.000, 0.067] \\
\bottomrule
\end{tabular}
}
\caption{CIs for the pairwise difference in medians of GND across the years for NeurIPS. The interval is highlighted in bold if the difference is significant. Note that an interval being close to 0 indicates no practical difference in the medians.}
\label{tab:neurips-stat-GND}
\end{table}

\begin{table}[H]
\centering
\resizebox{.35\textwidth}{!}{%
\begin{tabular}{lll}
\toprule
 & ARR-2024 & ARR-2025 \\
\midrule
ARR-2022 & [-0.100, 0.057] & [-0.100, 0.050] \\
ARR-2024 & - & [-0.057, 0.057] \\
\bottomrule
\end{tabular}
}
\caption{CIs for the pairwise difference in medians of GND across the years for ARR. The interval is highlighted in bold if the difference is significant. Note that an interval being close to 0 indicates no practical difference in the medians.}
\label{tab:arr-stat-GND}
\end{table}

\clearpage
\subsubsection{Result for accepted and rejected papers of ICLR}
\begin{figure*}[!h]
    \centering
    \includegraphics[width=0.6\linewidth]{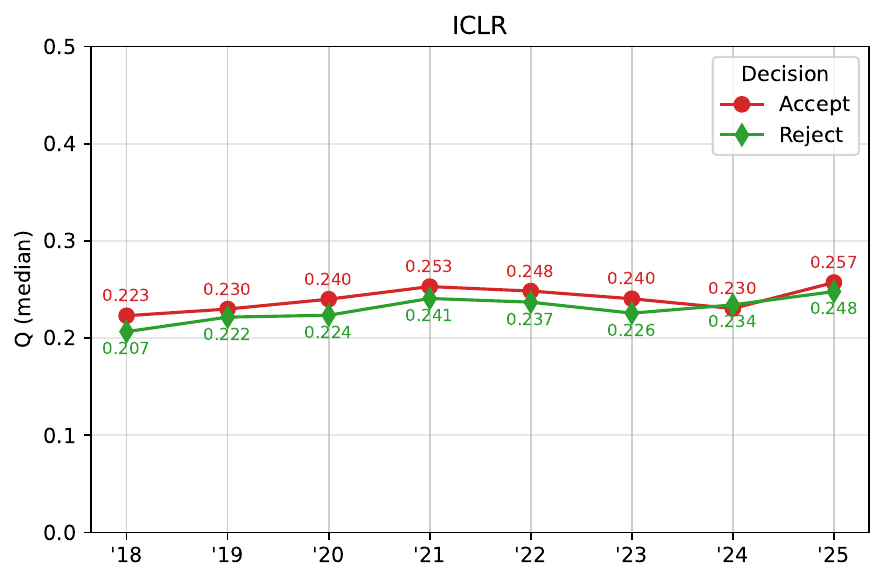}
    \caption{Q across time for accepted and rejected papers in ICLR.}
    \label{fig:Q-ICLR-acc-vs-rej}
\end{figure*}

\subsubsection{Results for Top and Bottom 10\% of reviews} \label{sec:app-worst-reviews}
\begin{figure*}[!h]
    \centering
    \includegraphics[width=\linewidth]{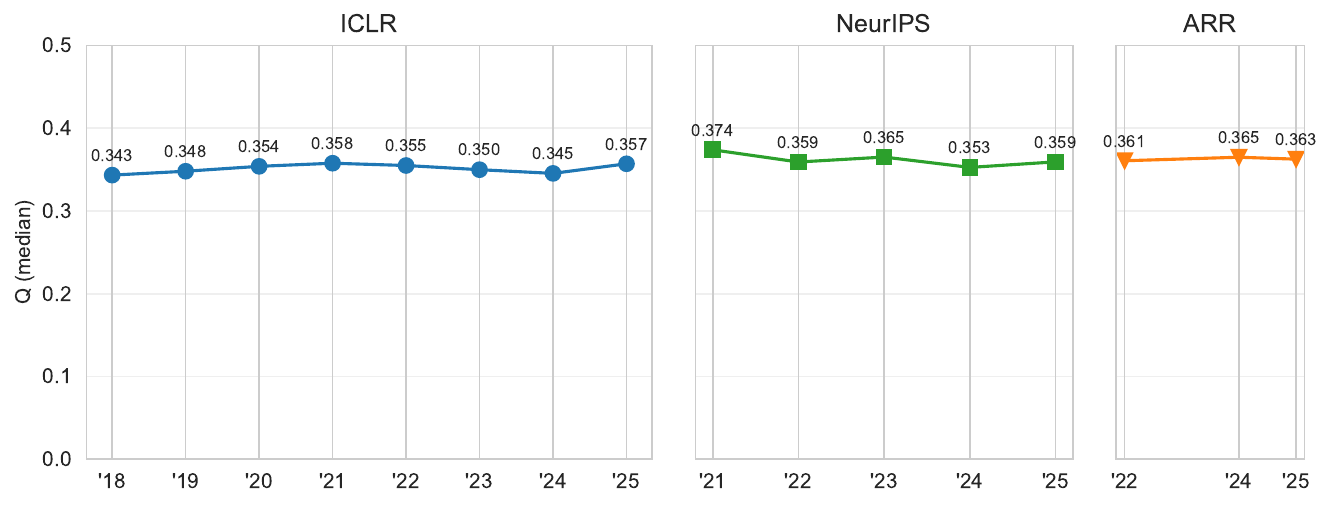}
    \caption{Q across venues and time -- top 10\% reviews.}
    \label{fig:Q-top}
\end{figure*}

\begin{figure*}[!h]
    \centering
    \includegraphics[width=\linewidth]{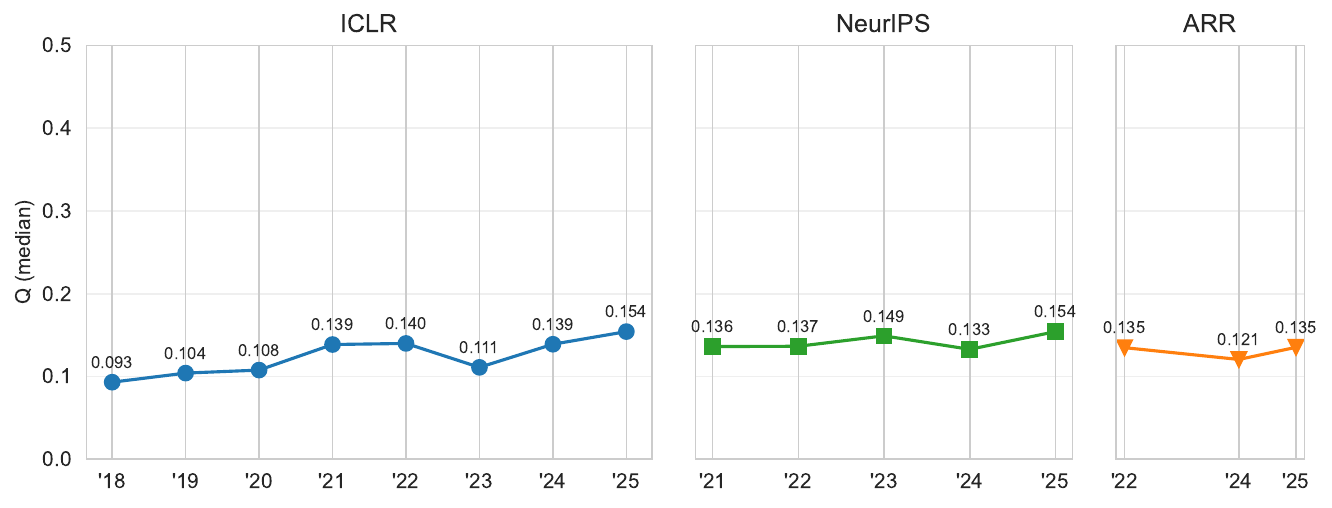}
    \caption{Q across venues and time -- bottom 10\% reviews.}
    \label{fig:Q-bottom}
\end{figure*}

\begin{table}[!h]
\centering
\resizebox{.95\textwidth}{!}{%
    \begin{tabular}{llllllll}
    \toprule
     & ICLR-2019 & ICLR-2020 & ICLR-2021 & ICLR-2022 & ICLR-2023 & ICLR-2024 & ICLR-2025 \\
    \midrule
    ICLR-2018 & [-0.014, 0.03] & [-0.008, 0.032] & \textbf{[0.027, 0.064]} & \textbf{[0.032, 0.064]} & [-0.001, 0.037] & \textbf{[0.028, 0.063]} & \textbf{[0.047, 0.076]} \\
    ICLR-2019 & - & [-0.018, 0.027] & \textbf{[0.019, 0.060]} & \textbf{[0.022, 0.060]} & [-0.013, 0.034] & \textbf{[0.019, 0.058]} & \textbf{[0.037, 0.073]} \\
    ICLR-2020 & - & - & \textbf{[0.015, 0.052]} & \textbf{[0.017, 0.055]} & [-0.015, 0.027] & \textbf{[0.015, 0.053]} & \textbf{[0.033, 0.068]} \\
    ICLR-2021 & - & - & - & [-0.013, 0.016] & [-\textbf{0.048, -0.009]} & [-0.016, 0.017] & \textbf{[0.002, 0.030]} \\
    ICLR-2022 & - & - & - & - & [-\textbf{0.048, -0.014]} & [-0.016, 0.014] & \textbf{[0.003, 0.025]} \\
    ICLR-2023 & - & - & - & - & - & \textbf{[0.011, 0.044]} & \textbf{[0.028, 0.059]} \\
    ICLR-2024 & - & - & - & - & - & - & \textbf{[0.002, 0.028]} \\
    \bottomrule
    \end{tabular}
}
\caption{CIs for the pairwise difference in medians across the years for 10\% lowest scoring reviews of ICLR. The interval is highlighted in bold if the difference is significant. Note that an interval being close to 0 indicates no practical difference in the medians.}
\label{tab:iclr-stat-bottom}
\end{table}

\begin{table}[!h]
\centering
\resizebox{.65\textwidth}{!}{%
    \begin{tabular}{lllll}
    \toprule
     & NeurIPS-2022 & NeurIPS-2023 & NeurIPS-2024 & NeurIPS-2025 \\
    \midrule
    NeurIPS-2021 & [-0.014, 0.018] & \textbf{[0.000, 0.030]} & [-0.016, 0.014] & \textbf{[0.005, 0.035]} \\
    NeurIPS-2022 & - & \textbf{[0.001, 0.024]} & [-0.017, 0.009] & \textbf{[0.005, 0.032]} \\
    NeurIPS-2023 & - & - & [-\textbf{0.028, -0.005]} & [-0.006, 0.017] \\
    NeurIPS-2024 & - & - & - & \textbf{[0.010, 0.034]} \\
    \bottomrule
    \end{tabular}
}
\caption{CIs for the pairwise difference in medians across the years for 10\% lowest scoring reviews of NeurIPS. The interval is highlighted in bold if the difference is significant. Note that an interval being close to 0 indicates no practical difference in the medians.}
\label{tab:neurips-stat-bottom}
\end{table}

\begin{table}[!h]
\centering
\resizebox{.35\textwidth}{!}{%
\begin{tabular}{lll}
\toprule
 & ARR-2024 & ARR-2025 \\
\midrule
ARR-2022 & [-0.033, 0.003] & [-0.018, 0.014] \\
ARR-2024 & - & [-0.005, 0.032] \\
\bottomrule
\end{tabular}
}
\caption{CIs for the pairwise difference in medians across the years for 10\% lowest scoring reviews of ARR. The interval is highlighted in bold if the difference is significant. Note that an interval being close to 0 indicates no practical difference in the medians.}
\label{tab:arr-stat-bottom}
\end{table}


\begin{table}[!h]
\centering
\resizebox{.95\textwidth}{!}{%
    \begin{tabular}{llllllll}
    \toprule
     & ICLR-2019 & ICLR-2020 & ICLR-2021 & ICLR-2022 & ICLR-2023 & ICLR-2024 & ICLR-2025 \\
    \midrule
    ICLR-2018 & [-0.014, 0.026] & [-0.005, 0.030] & [-0.005, 0.04] & [-0.003, 0.035] & [-0.008, 0.024] & [-0.013, 0.022] & [-0.002, 0.034] \\
    ICLR-2019 & - & [-0.015, 0.025] & [-0.017, 0.035] & [-0.011, 0.029] & [-0.018, 0.020] & [-0.020, 0.018] & [-0.010, 0.029] \\
    ICLR-2020 & - & - & [-0.015, 0.030] & [-0.014, 0.023] & [-0.020, 0.011] & [-0.023, 0.011] & [-0.013, 0.022] \\
    ICLR-2021 & - & - & - & [-0.028, 0.024] & [-0.032, 0.012] & [-0.034, 0.005] & [-0.027, 0.020] \\
    ICLR-2022 & - & - & - & - & [-0.026, 0.010] & [-0.031, 0.008] & [-0.022, 0.019] \\
    ICLR-2023 & - & - & - & - & - & [-0.021, 0.013] & [-0.011, 0.026] \\
    ICLR-2024 & - & - & - & - & - & - & [-0.003, 0.029] \\
    \bottomrule
    \end{tabular}
}
\caption{CIs for the pairwise difference in medians across the years for 10\% highest scoring reviews of ICLR. The interval is highlighted in bold if the difference is significant. Note that an interval being close to 0 indicates no practical difference in the medians.}
\label{tab:iclr-stat-top}
\end{table}

\begin{table}[!h]
\centering
\resizebox{.65\textwidth}{!}{%
    \begin{tabular}{lllll}
    \toprule
     & NeurIPS-2022 & NeurIPS-2023 & NeurIPS-2024 & NeurIPS-2025 \\
    \midrule
    NeurIPS-2021 & [-0.030, 0.010] & [-0.031, 0.012] & \textbf{[-0.035, -0.000]} & [-0.026, 0.002] \\
    NeurIPS-2022 & - & [-0.021, 0.028] & [-0.027, 0.017] & [-0.023, 0.016] \\
    NeurIPS-2023 & - & - & [-0.033, 0.018] & [-0.024, 0.022] \\
    NeurIPS-2024 & - & - & - & [-0.018, 0.022] \\
    \bottomrule
    \end{tabular}
}
\caption{CIs for the pairwise difference in medians across the years for 10\% highest scoring reviews of NeurIPS. The interval is highlighted in bold if the difference is significant. Note that an interval being close to 0 indicates no practical difference in the medians.}
\label{tab:neurips-stat-top}
\end{table}

\begin{table}[!h]
\centering
\resizebox{.35\textwidth}{!}{%
    \begin{tabular}{lll}
    \toprule
     & ARR-2024 & ARR-2025 \\
    \midrule
    ARR-2022 & [-0.016, 0.021] & [-0.016, 0.015] \\
    ARR-2024 & - & [-0.017, 0.010] \\
    \bottomrule
    \end{tabular}
}
\caption{CIs for the pairwise difference in medians across the years for 10\% highest scoring reviews of ARR. The interval is highlighted in bold if the difference is significant. Note that an interval being close to 0 indicates no practical difference in the medians.}
\label{tab:arr-stat-top}
\end{table}
    
\clearpage

\twocolumn
\subsection{Review itemization evaluation details}
\label{sec:a:itemeval}

\subsubsection{Guidelines}
\begin{figure*}[!h]
    \centering
    \includegraphics[width=\linewidth]{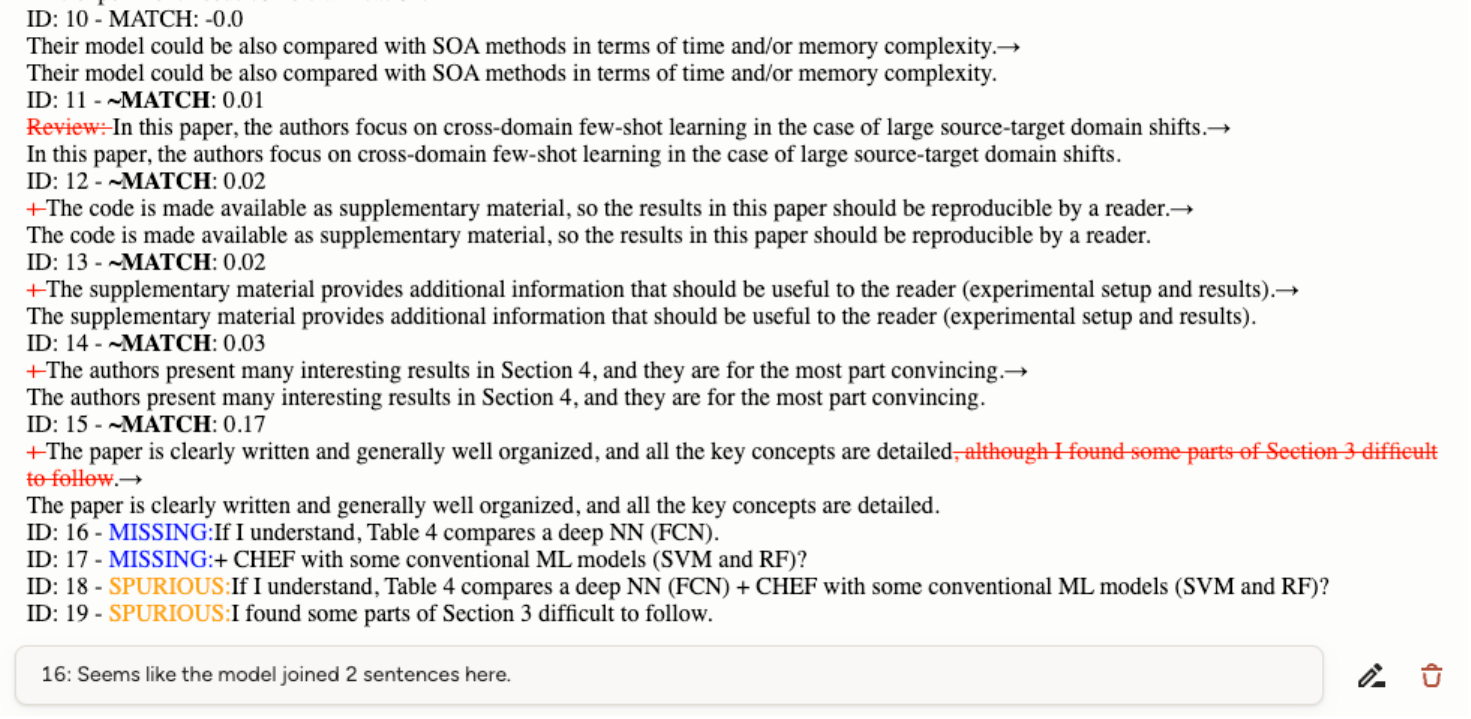}
    \caption{Review itemization evaluation interface: hallucinations and omissions.}
    \label{fig:seg-eval-1}
\end{figure*}

\begin{figure*}[!h]
    \centering
    \includegraphics[width=\linewidth]{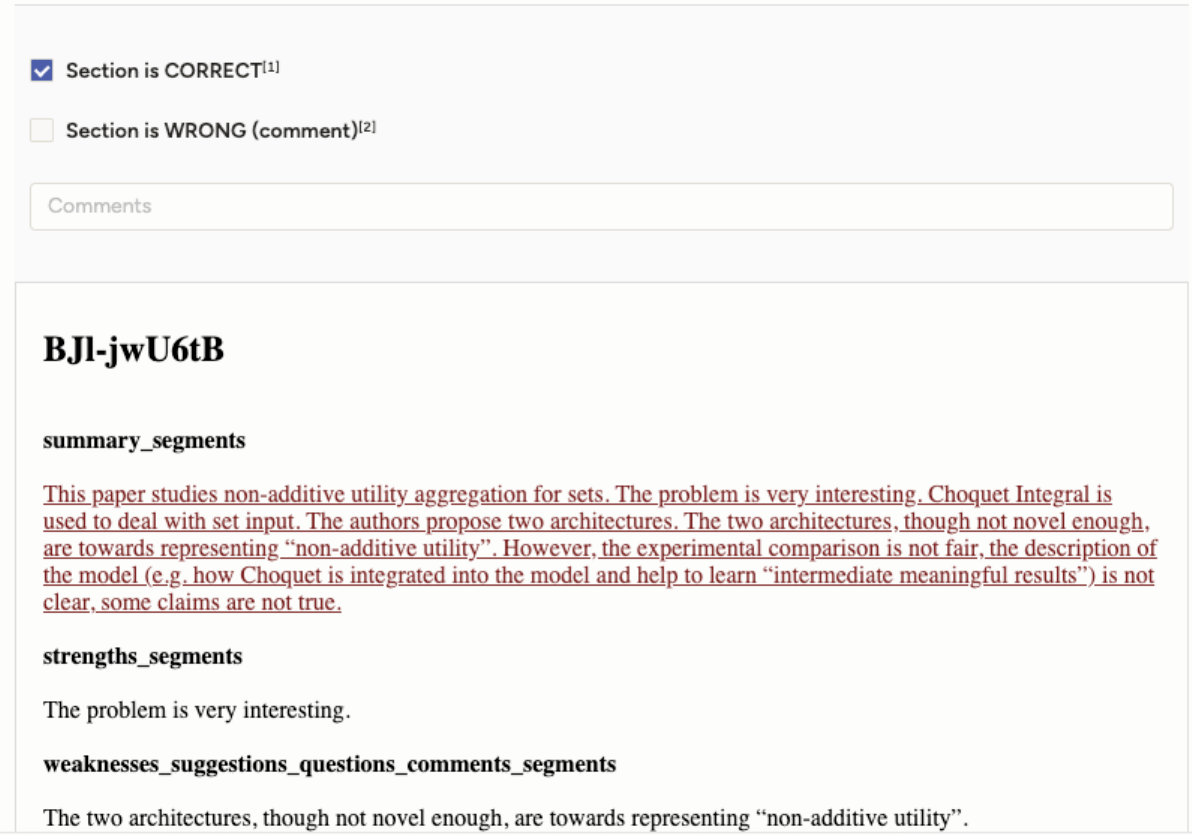}
    \caption{Review itemization evaluation interface: item-to-section assignment.}
    \label{fig:seg-eval-2}
\end{figure*}

\begin{figure*}[!h]
    \centering
    \includegraphics[width=\linewidth]{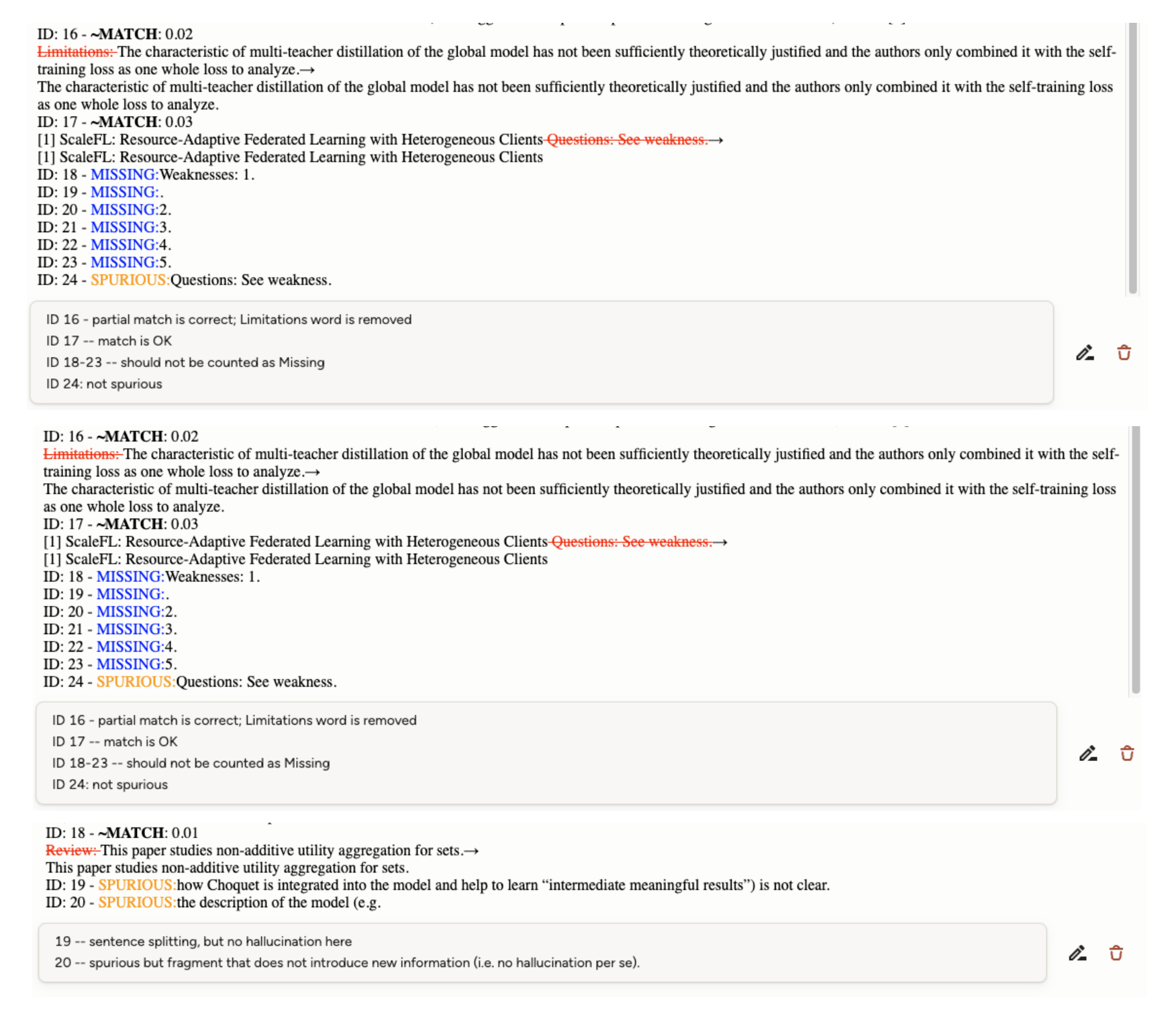}
    \caption{Examples of mismatches in review itemization, along with human annotator interpretations for hallucinations and omissions (Section \ref{sec:itemeval}).}
    \label{fig:seg-eval-3}
\end{figure*}

\begin{figure*}[!h]
    \centering
    \includegraphics[width=\linewidth]{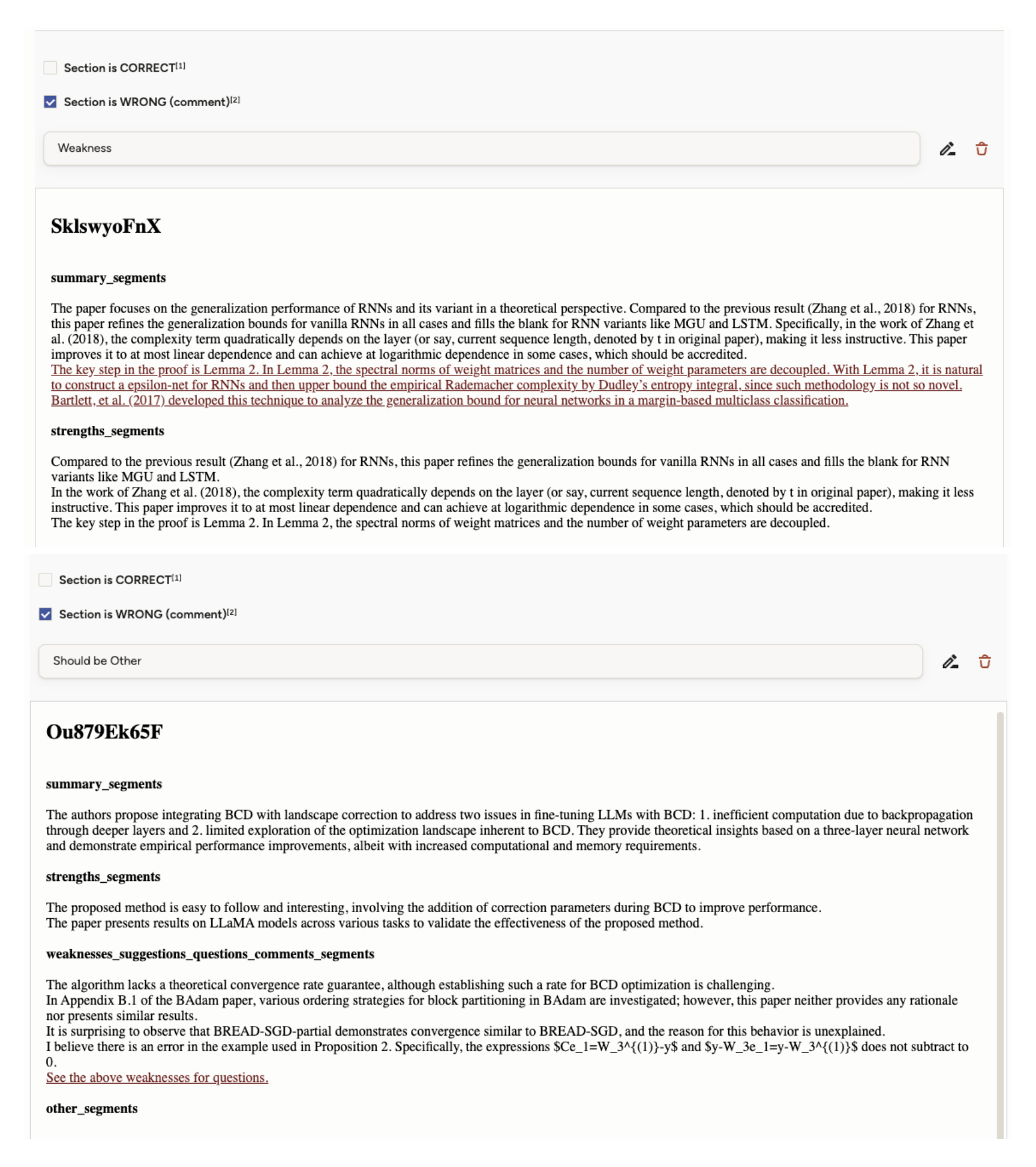}
    \caption{Examples of errors in item-to-section assignment (Section \ref{sec:itemeval}).}
    \label{fig:seg-eval-4}
\end{figure*}

The annotators were given a collection of semantically matched pairs of sentences from the original and itemized review, with the differences highlighted in red (Figure \ref{fig:seg-eval-1}). In addition, the sentences in the original review missing in the itemized review (potential omission) and the sentences in the itemized review not present in the original review (potential hallucination) were shown. The annotators were asked to judge if the itemization process introduced hallucinated content or omitted content from the original review by reviewing the partially matching, missing and spurious sentences for correctness, and to provide comments for each non-matching item. Errors due to splitting of sentences, omission of headers and minor paraphrases were ignored. For the evaluation of sections, the annotators were asked if the item was correctly assigned to a given section (Figure \ref{fig:seg-eval-2}). Further examples of mismatching in review itemization and section assignment are given in Figures \ref{fig:seg-eval-3} and \ref{fig:seg-eval-4}.

\onecolumn
\subsubsection{Interfaces and error examples}

\clearpage

\clearpage

\twocolumn
\subsection{Human rating validation study}
\label{subsec:app-humeval}

Since review quality judgment is prone to bias and \citet{rotrev} do not report inter-annotator agreement (IAA), to validate the review quality scores, we initially re-annotated a subset of 30 reviews from their 100 reviews with two expert annotators (one PhD-level, one faculty-level NLP) following the original guidelines from \cite{rotrev}. As shown in Table \ref{tab:agg_cikm}, we observed an average Krippendorff’s $\alpha$ of 0.18 for the individual 12 aspects used by \citet{rotrev} and 0.46 for overall quality, reflecting low and moderate IAA, respectively. This suggests that the original guidelines might be insufficient for independent deployment, and require task-specific instruction and training.

To address this, we developed a new guideline covering six dimensions of review quality (Section \ref{sec:app-guidelines}). Two annotators drafted the guidelines and subsequently refined them by labeling the same 30 reviews and discussing ambiguities. A third annotator (PhD student in NLP) was trained and completed a trial annotation of 3 reviews and labeled the same set of 30 reviews, yielding three annotations per review. On a subset of 30 reviews, we obtain an average IAA Krippendorff’s $\alpha$ of 0.55 for Substantiveness, Actionability, and Grounding (see Table \ref{tab:agg_ours}). Because the annotations for the remaining three aspects exhibit skewed distributions where the majority label accounts for over 90\% of cases, Krippendorff’s $\alpha$ for these aspects can be misleading~\citep{Jeni2013-yl, Zhao2022-xu}. Therefore, we measured IAA with Quadratic Weighted Agreement, yielding a pairwise average value of 0.97 among the three annotators for Stance, Objectivity and Readability, indicating strong agreement for the remaining three aspects.

Following this, the remaining 70 reviews were annotated in a single-annotator manner, and the overall review quality scores were computed by averaging across the six dimensions. On the subset of 30 reviews, correlations between overall scores and $Q$ across annotators were 0.5, 0.6 and 0.8 respectively (Spearman). To estimate an overall correlation across all 100 reviews, we sampled one annotation per item from the triply labeled subset and combined it with the 70 singly labeled reviews. Averaged over five independent runs, this yielded a correlation of 0.5, indicating moderate-strong alignment, providing further validation for our Q score.

\begin{table}[!h]
\centering
\resizebox{.35\textwidth}{!}{%
    \begin{tabular}{lr}
    \toprule
    Aspect & $\alpha$ \\
    \midrule
    Comprehensiveness & 0.179 \\
    Usage of Technical Terms & -0.118 \\
    Factuality & -0.074 \\
    Sentiment Polarity & 0.267 \\
    Politeness & -0.116 \\
    Vagueness & 0.156 \\
    Objectivity & 0.075 \\
    Fairness & -0.321 \\
    Actionability & 0.248 \\
    Constructiveness & 0.186 \\
    Relevance Alignment & -0.266 \\
    Clarity and Readability & -0.187 \\
    Overall Quality & 0.461 \\
    \bottomrule
    \end{tabular}
}
\caption{Krippendorff's $\alpha$ for aspects as defined by \citet{rotrev} for 3 annotators.}
\label{tab:agg_cikm}
\end{table}


\begin{table}[!h]
\centering
\resizebox{.35\textwidth}{!}{%
    \begin{tabular}{lrr}
    \toprule
    Aspect & $\alpha$ & QWA \\ 
    \midrule
    Substantiveness & 0.615 & - \\
    Actionability & 0.531 & - \\
    Grounding & 0.493 & - \\
    Stance & - & 0.953 \\
    Objectivity & - & 0.988 \\
    Readability & - & 0.982 \\ 
    \bottomrule
    \end{tabular}
}
\caption{Krippendorff's $\alpha$ and Quadratic Weighted Agreement (QWA) for each aspect defined in our guidelines.}
\label{tab:agg_ours}
\end{table}

\onecolumn
\subsection{Guidelines for assessing review quality} \label{sec:app-guidelines}
\begin{lstlisting}[breaklines=true, basicstyle=\tiny\ttfamily, frame=single, breakindent=0pt, columns=fullflexible]
# Actionability
A review must contain comments that are actionable for the authors. We consider two dimensions for judging each comment:
Is the action stated directly, or does the author need to infer it?
Explicit: The comment provides direct or apparent actions or suggestions. Authors can immediately identify the modifications needed for their draft. Clarification questions are considered explicit if they imply a direct action.
Implicit: The comment suggests actions indirectly. This can be in the form of questions that need to be addressed or missing parts that need to be added. Actions are not explicitly stated, but authors can infer what needs to be done after reading the comment.
After identifying the action, is it clear how to apply it, or is it vague?
Concrete: After identifying the action, the authors clearly understand what needs to be done and how to apply the suggested changes.
Vague: Even after identifying the action, the authors still do not know how to carry it out.
Focus only on Weaknesses +  Questions sections.

5 (Explicit + Concrete)
What will happen if you use the evaluation metric X instead of Y?

4 (Implicit + Concrete)
There are some very relevant baselines like X and Y that other people have been comparing their results to.

3 (Explicit + Vague)
You should address the lack of technical novelty in this paper.

2 (Implict + Vague)
It is not clear if this trend holds across different model architectures.

1 (Unactionable)
The idea of using positional encodings (PEs) for GNNs on molecular graph regression is not new.

Aggregation at review level is performed by identifying the applicable comments, averaging their individual scores and rounding up for the final score.

# Grounding
A good review must have specific comments that tell the authors clearly which part of the paper needs revision and why. We consider two dimensions:
What part of the paper does the comment address?
Weak grounded: The authors cannot precisely identify the section being addressed, but they have some hint or an educated guess.
Fully grounded: The authors can accurately determine which part of the paper is being referenced. This can be achieved through:
Explicit mentions of sections, tables, figures, etc.
Discussion of a unique aspect of the paper that allows the authors to infer the reference.
General comments that, while not mentioning specific parts, make it clear which sections are being addressed.
Not grounded
What is wrong with this part?
Measures the level of detail provided in identifying what is wrong or missing in the referenced part of the paper. If the comment refers to external work, specificity also evaluates whether concrete examples are provided. We consider "Specific", "Under-specific", "Not-Specific".
Focus only on Weaknesses +  Questions sections.

5 (Fully grounded + Specific)
The differences in results in Table 2 are very small that make the interpretation of results rather difficult. Furthermore, it is then unclear which proposed methods are really effective.

4 (Fully grounded + Under-specific)
In Figure 7, the results and supplemental video results show that SurfGAN seems out of place.

3 (Weakly Grounded + Specific)
Some figures need their captions to be more precise and to define all variables used in the figure.

2 (Weakly Grounded + Not-Specific)
For many of the datasets tested, the improvement over other approaches or even the general adversarial approach is marginal.

1 (Not Grounded)
The paper discusses a hot topic in the field now. However, one major drawback to this draft is that the analysis is poor.

Aggregation is again done by averaging individual scores and rounding up.

# Substantiveness
Measures the degree to which different elements of the paper are discussed in the review. Provide a score of 5 if all dimensions are discussed sufficiently. Note that we do not care about the discussion being specific here. We consider the following dimensions:
Contributions and Novelty
Methodology
Evaluation
Presentation, writing, clarity

Focus on the whole review.

5 (All aspects are discussed sufficiently)
Summary:
This paper proposes Data Prediction Denoising Models (DPDM)s, a new class of generative models that incorporates a sequence of strong denoisers for data generation. It is argued that conventional diffusion models embody weak denoisers, which in turn requires a high number of steps at inference time for generation, reducing the overall efficiency. To alleviate this challenge, DPDM training uses a teacher DM for initialization, and minimizes probability divergences between denoiser-recovered data distributions and the ground truth data distribution. In addition, a sampler suitable for DPDMs is presented. It is shown that DPDM can attain strong performance on CIFAR-10, and ImageNet64x64 with only a few number of sampling steps.

Strengths:
- The proposed method is novel, and the technical contribution is strong. DPDMs provide a new framework for generative modeling which is different from diffusion models, although it draws many inspirations from diffusion models.
- The experimental results are convincing where DPDMs outperform many recent baselines, illustrating their capability on data generation using a small number of NFEs.

Weaknesses:
- The scope of the experiments section is limited where the results are presented on low-resolution datasets such as CIFAR-10, and ImageNet 64x64. The experiments would benefit from demonstrations with high-resolution datasets to illustrate the generalizability of the method. 
- Although the focus of the paper is on inference efficiency, inference metrics are not provided. The comparison is made in terms of NFEs which is an important aspect. However, metrics such as inference memory, seconds per iteration at inference, and overall inference time until convergence should be provided for a valid comparison.
- Although the paper is well-written, there are typographical issues such as duplicate references, (Song et al., 2020b and Song et al., 2020c), duplicate paragraphs (Training Efficiency and GPU-memory Cost in main and appendix) and other issues which should be fixed.

Questions:
- How does inference memory and time compare with state-of-the-art baselines?
- What does $\tilde{x}$ stand for in Equation 1? I don't think it has been defined anywhere.

4 (3 aspects are discussed sufficiently with 1 aspect being discussed briefly)
Summary:
This paper proposes Data Prediction Denoising Models (DPDM)s, a new class of generative models that incorporates a sequence of strong denoisers for data generation. It is argued that conventional diffusion models embody weak denoisers, which in turn requires a high number of steps at inference time for generation, reducing the overall efficiency. To alleviate this challenge, DPDM training uses a teacher DM for initialization, and minimizes probability divergences between denoiser-recovered data distributions and the ground truth data distribution. In addition, a sampler suitable for DPDMs is presented. It is shown that DPDM can attain strong performance on CIFAR-10, and ImageNet64x64 with only a few number of sampling steps.

Strengths:
- The proposed method is novel, and the technical contribution is strong. DPDMs provide a new framework for generative modeling which is different from diffusion models, although it draws many inspirations from diffusion models.
- The experimental results are convincing where DPDMs outperform many recent baselines, illustrating their capability on data generation using a small number of NFEs.

Weaknesses:
- The scope of the experiments section is limited where the results are presented on low-resolution datasets such as CIFAR-10, and ImageNet 64x64. The experiments would benefit from demonstrations with high-resolution datasets to illustrate the generalizability of the method. 
- Although the focus of the paper is on inference efficiency, inference metrics are not provided. The comparison is made in terms of NFEs which is an important aspect. However, metrics such as inference memory, seconds per iteration at inference, and overall inference time until convergence should be provided for a valid comparison.
- Although the paper is well-written, there are typographical issues which should be fixed.

Questions:
- How does inference memory and time compare with state-of-the-art baselines?
- What does $\tilde{x}$ stand for in Equation 1? I don't think it has been defined anywhere.

3 (2 aspects are discussed sufficiently with other aspects being discussed briefly or 1 aspect being omitted)
Summary:
This paper proposes Data Prediction Denoising Models (DPDM)s, a new class of generative models that incorporates a sequence of strong denoisers for data generation. It is argued that conventional diffusion models embody weak denoisers, which in turn requires a high number of steps at inference time for generation, reducing the overall efficiency. To alleviate this challenge, DPDM training uses a teacher DM for initialization, and minimizes probability divergences between denoiser-recovered data distributions and the ground truth data distribution. In addition, a sampler suitable for DPDMs is presented. It is shown that DPDM can attain strong performance on CIFAR-10, and ImageNet64x64 with only a few number of sampling steps.

Strengths:
- The proposed method is novel, and the technical contribution is strong. DPDMs provide a new framework for generative modeling which is different from diffusion models, although it draws many inspirations from diffusion models.
- The experimental results are convincing where DPDMs outperform many recent baselines, illustrating their capability on data generation using a small number of NFEs.

Weaknesses:
- The scope of the experiments section is limited.
- Although the focus of the paper is on inference efficiency, inference metrics are not provided.

Questions:
- How does inference memory and time compare with state-of-the-art baselines?
- What does $\tilde{x}$ stand for in Equation 1? I don't think it has been defined anywhere.

2 (2 aspects are discussed while other aspects are omitted)
Summary:
This paper proposes Data Prediction Denoising Models (DPDM)s, a new class of generative models that incorporates a sequence of strong denoisers for data generation. It is argued that conventional diffusion models embody weak denoisers, which in turn requires a high number of steps at inference time for generation, reducing the overall efficiency. To alleviate this challenge, DPDM training uses a teacher DM for initialization, and minimizes probability divergences between denoiser-recovered data distributions and the ground truth data distribution. In addition, a sampler suitable for DPDMs is presented. It is shown that DPDM can attain strong performance on CIFAR-10, and ImageNet64x64 with only a few number of sampling steps.

Strengths:
- The proposed method is novel, and the technical contribution is strong. DPDMs provide a new framework for generative modeling which is different from diffusion models, although it draws many inspirations from diffusion models.
- The experimental results are convincing where DPDMs outperform many recent baselines, illustrating their capability on data generation using a small number of NFEs.

Weaknesses:
- Evaluation has to be improved.

Questions:
- How does inference memory and time compare with state-of-the-art baselines?
- What does $\tilde{x}$ stand for in Equation 1? I don't think it has been defined anywhere.

1 (One or no aspects are discussed)
Summary:
This paper proposes Data Prediction Denoising Models (DPDM)s, a new class of generative models that incorporates a sequence of strong denoisers for data generation.

Strengths:
- The proposed method is novel.
- The experimental results are convincing.

Weaknesses:
- Evaluation has to be improved.

# Stance
A review must be written in such a way that its stance regarding the acceptance/rejection of the paper is clear and well supported. We consider two dimensions:
Is the stance clear?
Clear: Explicitly mentioned or easily inferred from arguments (requires arguments).
Unclear: Not explicitly mentioned and cannot be inferred from arguments (arguments may or may not present).
Are there arguments present for justifying the stance? Arguments could be in strengths, weaknesses or other parts of the review. Here we do not care about specificity of the arguments.
Present and well-justified: Arguments present and stance is well-justified.
Present and not well-justified: Arguments present but do not justify stance in a clear way.
Present and almost non-existent: Arguments present but are mentioned briefly.

Focus on the whole review.

4 (Clear + Present and well-justified)
Despite the issues with presentation and lack of baselines, the paper introduces a novel method for circuit detection and provides exhaustive evaluation results, therefore being impactful.

3 (Unclear + Present and not well-justified)
The paper introduces a novel method for circuit detection and provides exhaustive evaluation results, but has multiple issues with presentation and lacks comparison with baselines.

2 (Clear + Present and almost non-existent)
The paper is novel, so I recommend a strong accept.

1 (Unclear + Present and almost non-existent)
-

# Objectivity
A review must provide a balanced judgement of the paper and not focus narrowly only on strengths or on weaknesses. Provide a score of 5 if both strengths and weaknesses are judged fairly and a score of 1 if the review exclusively focuses either on the strengths or on the weaknesses. 
Note that it is very hard to judge objectivity, so here we are trying to give a low score if the review is obviously not objective.
Focus only on Strengths and Weaknesses + Questions sections.

# Readability
A review must be easy to understand and follow. We consider two dimensions here: 
Structure: Is the information structured and segregated as required?
Well structured: Review contains clear dedicated sections with bullet points inside each section.
Moderately structured: Review contains different sections/paragraphs but no bullet points.
Badly structured: Information is not organized well into sections/paragraphs making it harder to follow.
Grammar: Is the grammar confusing or easy to understand?
Understandable grammar: The grammar used is easily understandable possibly even in the presence of typos.
Confusing grammar: The review text is ambiguous due to grammatical mistakes and awkward phrases.

P.S.: It is possible that there will be no points within a section if it discusses just one thing.

Focus on the whole review.

5 (Well structured + Understandable grammar)

4 (Moderately structured + Understandable grammar)

3 (Well structured + Confusing grammar)

2 (Moderately structured + Confusing grammar or Badly structured + Understandable grammar)

1 (Badly structured + Confusing grammar)
\end{lstlisting}

\clearpage
\subsection{Prompt for itemization} \label{sec:app-prompt}

\begin{lstlisting}[breaklines=true, basicstyle=\tiny\ttfamily, frame=single, breakindent=0pt, columns=fullflexible]
# Task Description
You are given a review from a conference which may have a structure of its own or be completely unstructured. Your objective is to arrange the review segments into 4 sections to obtain the review in a structured format. The 4 sections are: Summary, Strengths, Weaknesses/Suggestions/Questions/Comments and Other. Strengths and Weaknesses/Suggestions/Questions/Comments need to have separated points while Summary does not. Each point should be focused on one claim and have all the context mentioned in the review for it. You may have to combine different segments which are not consecutively placed in the original review into one point to achieve this. On the other hand, it is also possible that a single text discussing multiple ideas may need to be split into different points. When doing this, make sure to repeat any text necessary so that the context is not lost. Any additional information which does not fit into Summary, Strengths and Weaknesses/Suggestions/Questions/Comments sections should be placed in the Other section. You are provided with 4 examples on how to do this task. Follow the examples closely while structuring the review.

# Rules
1. Avoid paraphrasing unless absolutely needed for coherence and do not change the meaning of any part of the review. 
2. Make sure to minimize adding new content or discarding any content from the given review. 
3. If details for a section is missing in the original review, you can leave that section empty in the output.
4. Do not correct any grammatical or spelling mistakes in the review.
5. The final output should in JSON format containing a list of strings for each section, with each string for Strengths and Weaknesses/Suggestions/Questions/Comments sections corresponding to one point.

# Examples
## Example Review 1 Input
Summary:
The authors introduce Conditional Activation Steering (CAST) and condition vectors. They show that refusal behaviors can be invoked conditionally on the context of the prompt allowing for conditional steering. They test this across several language models up to size 8B and show that their method has fewer false positive refusals on harmless prompts while still maintaining high refusal rates on harmful prompts, demonstrating the effectiveness of the conditional.

Strengths:
1. The method of steering LLM conditionally on the context of the prompt is novel and an important contribution towards practical implementations of activation steering.
2. The ability to chain conditionals is an interesting contribution.
3. The paper is relatively thorough in its test of models within a certain class O(8B).

Weaknesses:
1. All the tested models have less than or equal to 8B parameters. Testing on larger models would help improve the robustness and confidence in the results
2. (Minor) The harmless/harmful refusals are not tested against enough real-world inputs, like jailbreaks or multi-turn conversations.
3. (Minor) There is no limitations or future work section.

Questions:
The paper is generally well written and was pleasant and interesting to read. The possibility for conditional steering is exciting with many practical implications.

### Minor
The paper could be improved through some careful revisions to the figures and layouts.
* Figure 1 is presented too early; its full description is on page 6 while it appears at the top of page 2. Despite being referenced on page 1, the paper would flow better if Figure 1 were closer to page 6.
* All of the T-SNE plots should consider a different color scheme. Its very difficult to distinguish the Alpaca vs Sorry-bench dots, especially against the background of a similar color.
* Figure 8c, several pieces of text are too small to easily read
* Figure 9, the label "(c)" is not placed in the top left corner like the previous figures. The markers are difficult to see (e.g. the start marker).


### Out-of-scope improvements
While the following improvements would substantially increase the value of the paper, the reviewer recognizes that they can be designated to follow up work and may be out-of-scope for the current paper.
* The methods could be tested with models with > 8B parameters
* The methods could be tested against known jailbreaks (e.g. does the conditional vector for harmfulness still activation on a zero-shot or 1-shot jailbreak?).

## Example Review 1 Output
{
  "summary_segments": [
    "The authors introduce Conditional Activation Steering (CAST) and condition vectors. They show that refusal behaviors can be invoked conditionally on the context of the prompt allowing for conditional steering. They test this across several language models up to size 8B and show that their method has fewer false positive refusals on harmless prompts while still maintaining high refusal rates on harmful prompts, demonstrating the effectiveness of the conditional."
  ],
  "strengths_segments": [
    "The method of steering LLM conditionally on the context of the prompt is novel and an important contribution towards practical implementations of activation steering.",
    "The ability to chain conditionals is an interesting contribution.",
    "The paper is relatively thorough in its test of models within a certain class O(8B).",
    "The paper is generally well written and was pleasant and interesting to read. The possibility for conditional steering is exciting with many practical implications."
  ],
  "weaknesses_suggestions_questions_comments_segments": [
    "All the tested models have less than or equal to 8B parameters. Testing on larger models would help improve the robustness and confidence in the results",
    "(Minor) The harmless/harmful refusals are not tested against enough real-world inputs, like jailbreaks or multi-turn conversations.",
    "(Minor) There is no limitations or future work section.",
    "The paper could be improved through some careful revisions to the figures and layouts.
* Figure 1 is presented too early; its full description is on page 6 while it appears at the top of page 2. Despite being referenced on page 1, the paper would flow better if Figure 1 were closer to page 6.",
    "The paper could be improved through some careful revisions to the figures and layouts.
* All of the T-SNE plots should consider a different color scheme. Its very difficult to distinguish the Alpaca vs Sorry-bench dots, especially against the background of a similar color.",
    "The paper could be improved through some careful revisions to the figures and layouts.
* Figure 8c, several pieces of text are too small to easily read",
    "The paper could be improved through some careful revisions to the figures and layouts.
* Figure 9, the label "(c)" is not placed in the top left corner like the previous figures. The markers are difficult to see (e.g. the start marker).",
    "While the following improvements would substantially increase the value of the paper, the reviewer recognizes that they can be designated to follow up work and may be out-of-scope for the current paper.
* The methods could be tested with models with > 8B parameters",
    "While the following improvements would substantially increase the value of the paper, the reviewer recognizes that they can be designated to follow up work and may be out-of-scope for the current paper.
* The methods could be tested against known jailbreaks (e.g. does the conditional vector for harmfulness still activation on a zero-shot or 1-shot jailbreak?)."
  ],
  "other_segments": []
}


## Example Review 2 Input
Summary:
This paper explores interleaved multimodal conversations with LLMs. The authors first construct a multimodal conversation instruction tuning dataset with text-only GPT-4 and image descriptions. Then, multi-turn interleaved multimodal (dubbed MIM) instruction tuning based on GILL is used to train the multimodal LLMs to learn which segment of text information should be used for diffusion model image synthesis (which also defines when to generate images). Experiments demonstrate promising results.

Strengths:
- The targeted problem of enabling multimodal LLMs to generate images for a multimodal multi-turn conversation is trending, interesting, and important. Great application potentials can be induced in both the research community and the industry.
- The proposed dataset would be very useful to the community if it is open-sourced. The construction involves human-in-the-loop refinement, which is good since the data from the internet is extremely noisy.

Weaknesses:
- **Major concern 1.** My first major concern lies in the novelty of the proposed textual exchange method for enabling LLMs to generate images. As far as I know, learning which segment of texts to use as the text inputs of text-to-image models was first proposed by Divter [1] (not cited or discussed). Divter proposes to learn the textural inputs by using a special token and constructed template. However, leveraging LLMs and diffusion models is different, but the contribution is still limited in this case. The authors are required to clarify the similarities and differences with Divter. On the other hand, using only textual conditions to generate images has its limitations when considering very long-context-based (e.g., interleaved documents) image generation or image-conditioned image generation (e.g., image2image translation, image edition, etc).
- **Major concern 2.** My second major concern lies in the experimental evaluations. i) In Table 3, more commonly used NLP benchmarks like MMLU, HellaSwag, and WinoGrande should be conducted. In Table 5, what about the most commonly used metric FID results? In Table 6, I wonder about the zero-shot results of TextBind on VQAv2 and MM-Vet. ii) No ablation studies or in-depth discussions are presented. For example, what if no human-in-the-loop is used during dataset construction? What are the failure cases of TextBind, and why? What emerging properties could be explored by TextBind?
- **Technical contribution.** The proposed method is simple, but the technical contribution is limited. The proposed multimodal LLM architecture is mainly based on previous work GILL. The only difference is the textural information exchange method which is good but somewhat incremental. Besides, the topic-aware image sampling is quite similar to the dataset constructed by PALI-X [2]. PALI-X constructs its own interleaved dataset Episodic WebLI by grouping image-text pairs.
- **Unclearly supported claims.** The authors claim that the current multimodal instruction tuning methods lead to limited performance in open-world scenarios. However, there is a lack of analysis of TextBind's superiority in it. Besides, annotation-free may be overclaimed since human-in-the-loop definitely requires non-trivial annotation efforts.

[1] Multimodal Dialogue Response Generation. In ACL 2022.
[2] PaLI-X: On Scaling up a Multilingual Vision and Language Model.

Questions:
- Why is the method called TextBind? Assuming the authors are trying to analog to ImageBind [3]. However, the core spirit of multimodal binding in the same embedding space as a modality-agnostic multimodal encoder is very different from this paper. I am quite confused about this. A name that better summarizes the work's idea is better than using one similar to an existing work while not very suitable.
- Will the curated dataset be released to the public?
- The dataset is constructed by using image descriptions with GPT-4, similar to LLaVA [4] and ChatCaptioner [5] (not cited). I wonder how is the hallucination problem of data and models in such progress since CLIP filtering can not guarantee the avoidance of such an issue. For example, can authors provide some failure cases of the dataset and test the model's hallucination capability?
- Now we have GPT-4V, I am looking forward to the newly constructed dataset with GPT-4V (not required at this moment).
- Since Q-Former is used, which may compress the visual signals, I wonder about the OCR performance of TextBind. For example, zero-shot results on TextVQA?
- There are many concurrent works working in this direction. It would be good if these works were discussed in related work [6-10].
- Minor: When the abbreviation `MIM` first appears in the paper, there is no explanation of the meanings.

I am looking forward to the authors' response.

[3] ImageBind: One Embedding Space To Bind Them All. In CVPR 2023.
[4] Visual instruction tuning. In NeurIPS 2023.
[5] ChatGPT Asks, BLIP-2 Answers: Automatic Questioning Towards Enriched Visual Descriptions. arXiv 2023.
[6] Generative Pretraining in Multimodality. arXiv 2023.
[7] DreamLLM: Synergistic Multimodal Comprehension and Creation. arXiv 2023.
[8] MiniGPT-5: Interleaved Vision-and-Language Generation via Generative Vokens. arXiv 2023.
[9] NExT-GPT: Any-to-Any Multimodal LLM. arXiv 2023.
[10] Kosmos-G: Generating Images in Context with Multimodal Large Language Models. arXiv 2023.

## Example Review 2 Output
{
  "summary_segments": [
    "This paper explores interleaved multimodal conversations with LLMs. The authors first construct a multimodal conversation instruction tuning dataset with text-only GPT-4 and image descriptions. Then, multi-turn interleaved multimodal (dubbed MIM) instruction tuning based on GILL is used to train the multimodal LLMs to learn which segment of text information should be used for diffusion model image synthesis (which also defines when to generate images). Experiments demonstrate promising results."
  ],
  "strengths_segments": [
    "The targeted problem of enabling multimodal LLMs to generate images for a multimodal multi-turn conversation is trending, interesting, and important. Great application potentials can be induced in both the research community and the industry.",
    "The proposed dataset would be very useful to the community if it is open-sourced. The construction involves human-in-the-loop refinement, which is good since the data from the internet is extremely noisy."
  ],
  "weaknesses_suggestions_questions_comments_segments": [
    "**Major concern 1.** My first major concern lies in the novelty of the proposed textual exchange method for enabling LLMs to generate images. As far as I know, learning which segment of texts to use as the text inputs of text-to-image models was first proposed by Divter [1] (not cited or discussed). Divter proposes to learn the textural inputs by using a special token and constructed template. However, leveraging LLMs and diffusion models is different, but the contribution is still limited in this case. The authors are required to clarify the similarities and differences with Divter. On the other hand, using only textual conditions to generate images has its limitations when considering very long-context-based (e.g., interleaved documents) image generation or image-conditioned image generation (e.g., image2image translation, image edition, etc).",
    "**Major concern 2.** My second major concern lies in the experimental evaluations. i) In Table 3, more commonly used NLP benchmarks like MMLU, HellaSwag, and WinoGrande should be conducted. In Table 5, what about the most commonly used metric FID results? In Table 6, I wonder about the zero-shot results of TextBind on VQAv2 and MM-Vet. ii) No ablation studies or in-depth discussions are presented. For example, what if no human-in-the-loop is used during dataset construction? What are the failure cases of TextBind, and why? What emerging properties could be explored by TextBind?",
    "**Technical contribution.** The proposed method is simple, but the technical contribution is limited. The proposed multimodal LLM architecture is mainly based on previous work GILL. The only difference is the textural information exchange method which is good but somewhat incremental. Besides, the topic-aware image sampling is quite similar to the dataset constructed by PALI-X [2]. PALI-X constructs its own interleaved dataset Episodic WebLI by grouping image-text pairs.",
    "**Unclearly supported claims.** The authors claim that the current multimodal instruction tuning methods lead to limited performance in open-world scenarios. However, there is a lack of analysis of TextBind's superiority in it. Besides, annotation-free may be overclaimed since human-in-the-loop definitely requires non-trivial annotation efforts.",
    "Why is the method called TextBind? Assuming the authors are trying to analog to ImageBind [3]. However, the core spirit of multimodal binding in the same embedding space as a modality-agnostic multimodal encoder is very different from this paper. I am quite confused about this. A name that better summarizes the work's idea is better than using one similar to an existing work while not very suitable.",
    "Will the curated dataset be released to the public?",
    "The dataset is constructed by using image descriptions with GPT-4, similar to LLaVA [4] and ChatCaptioner [5] (not cited). I wonder how is the hallucination problem of data and models in such progress since CLIP filtering can not guarantee the avoidance of such an issue. For example, can authors provide some failure cases of the dataset and test the model's hallucination capability?",
    "Now we have GPT-4V, I am looking forward to the newly constructed dataset with GPT-4V (not required at this moment).",
    "Since Q-Former is used, which may compress the visual signals, I wonder about the OCR performance of TextBind. For example, zero-shot results on TextVQA?",
    "There are many concurrent works working in this direction. It would be good if these works were discussed in related work [6-10].",
    "Minor: When the abbreviation `MIM` first appears in the paper, there is no explanation of the meanings."
  ],
  "other_segments": [
    "[1] Multimodal Dialogue Response Generation. In ACL 2022.
[2] PaLI-X: On Scaling up a Multilingual Vision and Language Model.",
    "I am looking forward to the authors' response.",
    "[3] ImageBind: One Embedding Space To Bind Them All. In CVPR 2023.
[4] Visual instruction tuning. In NeurIPS 2023.
[5] ChatGPT Asks, BLIP-2 Answers: Automatic Questioning Towards Enriched Visual Descriptions. arXiv 2023.
[6] Generative Pretraining in Multimodality. arXiv 2023.
[7] DreamLLM: Synergistic Multimodal Comprehension and Creation. arXiv 2023.
[8] MiniGPT-5: Interleaved Vision-and-Language Generation via Generative Vokens. arXiv 2023.
[9] NExT-GPT: Any-to-Any Multimodal LLM. arXiv 2023.
[10] Kosmos-G: Generating Images in Context with Multimodal Large Language Models. arXiv 2023."
  ]
}


## Example Review 3 Input
Review:
The paper proposed a RNN with skip-connection (external memory) to past hidden states, this is a slightly different version of the TARDIS network. The authors experimented on PTB and a temporal action detection method.

Novelty:

I dont see a lot of novelty to the method. The authors proposed a method very similar to TARDIS, the difference seems to be that MMARNN does not use extra usage vectors for reading from previous memory, but this is not a fundamental difference between MMARNN and Tardis.

Shortcomings of the paper:

1. The experiments seem rather weak. The authors experimented on PTB and temporal action detection method. It is not clear why authors experimented with PTB, this is not a task with long-term dependencies, I do not see how this task (compared to many other tasks) can benefit from using external memory (especially when only 1 past hidden state is used

2. The model uses a single past hidden state, it is not clear to me why this is better than using a weighted sum of a few past hidden states, as many tasks requires long-term dependencies from multiple steps in the past. The authors should cite "Sparse attentive backtracking" (https://arxiv.org/abs/1809.03702) at NIPS 2018. SAB is very related in that it also propagate gradients to a few hidden states in the memory. The difference is that SAB used a few hidden states from the past/ memory instead of one; another difference is that it propagates gradients locally to the selected hidden states/ memory slots.

3. The paper only demonstrated experimental results on PTB and temporal action prediction. I think it would make the paper a lot stronger if the authors experimented with a variety of different tasks. Tasks that requires long term dependencies can really demonstrate the strength of the model (copy and adding tasks).

4. If the authors could run the model on copy and adding tasks, I would be curious to see if the model is picking the "correct" timestep in the memory / past.

post rebuttal: I feel that the authors have addressed some of my concerns, in particular, in terms of additional experimental results. I have raised the score to reflect this changes.

## Example Review 3 Output
{
  "summary_segments": [
    "The paper proposed a RNN with skip-connection (external memory) to past hidden states, this is a slightly different version of the TARDIS network. The authors experimented on PTB and a temporal action detection method."
  ],
  "strengths_segments": [],
  "weaknesses_suggestions_questions_comments_segments": [
    "I dont see a lot of novelty to the method. The authors proposed a method very similar to TARDIS, the difference seems to be that MMARNN does not use extra usage vectors for reading from previous memory, but this is not a fundamental difference between MMARNN and Tardis.",
    "The experiments seem rather weak. The authors experimented on PTB and temporal action detection method. It is not clear why authors experimented with PTB, this is not a task with long-term dependencies, I do not see how this task (compared to many other tasks) can benefit from using external memory (especially when only 1 past hidden state is used",
    "The model uses a single past hidden state, it is not clear to me why this is better than using a weighted sum of a few past hidden states, as many tasks requires long-term dependencies from multiple steps in the past. The authors should cite "Sparse attentive backtracking" (https://arxiv.org/abs/1809.03702) at NIPS 2018. SAB is very related in that it also propagate gradients to a few hidden states in the memory. The difference is that SAB used a few hidden states from the past/ memory instead of one; another difference is that it propagates gradients locally to the selected hidden states/ memory slots.",
    "The paper only demonstrated experimental results on PTB and temporal action prediction. I think it would make the paper a lot stronger if the authors experimented with a variety of different tasks. Tasks that requires long term dependencies can really demonstrate the strength of the model (copy and adding tasks).",
    "If the authors could run the model on copy and adding tasks, I would be curious to see if the model is picking the "correct" timestep in the memory / past."
  ],
  "other_segments": [
    "post rebuttal: I feel that the authors have addressed some of my concerns, in particular, in terms of additional experimental results. I have raised the score to reflect this changes."
  ]
}


## Example Review 4 Input
Summary:
This paper investigates the relationship between the batch-size and convergence speed for a broader class of nonconvex problems. Unfortunately, the results are not novel, the introduction is very short it isn not clear the main contribution of the paper, and the comparison with related work is quite brief without much in-depth discussion.

Strengths:
N/A

Weaknesses:
The submission is unfortunately not very strong. The main results (Theorem 1.1) shows that the effective noise level reduces with batch size, which seems trivial. Indeed, I cannot see much novelty and intuition provided by Theorem 1.1.

---------------------

The introduction is very short it isn not clear the main contribution of the paper, and the comparison with related work is quite brief without much in-depth discussion.

---------------------

The paper considers a narrow optimization problem, and the setting in Theorem 1.1 is also quite limited.

Therefore, I do recommend rejection.

Questions:
Please see above.

## Example Review 4 Output
{
  "summary_segments": [
    "This paper investigates the relationship between the batch-size and convergence speed for a broader class of nonconvex problems. Unfortunately, the results are not novel, the introduction is very short it isn not clear the main contribution of the paper, and the comparison with related work is quite brief without much in-depth discussion."
  ],
  "strengths_segments": [],
  "weaknesses_suggestions_questions_comments_segments": [
    "The submission is unfortunately not very strong. The main results (Theorem 1.1) shows that the effective noise level reduces with batch size, which seems trivial. Indeed, I cannot see much novelty and intuition provided by Theorem 1.1.",
    "The introduction is very short it isn not clear the main contribution of the paper, and the comparison with related work is quite brief without much in-depth discussion.",
    "The paper considers a narrow optimization problem, and the setting in Theorem 1.1 is also quite limited."
  ],
  "other_segments": [
    "Therefore, I do recommend rejection.",
    "Please see above."
  ]
}

# Review to be structured
{review}
\end{lstlisting}
\end{document}